\documentclass[letterpaper, 10 pt, conference]{ieeeconf}  

\IEEEoverridecommandlockouts                              
\usepackage{caption}
\usepackage{esvect}
\usepackage{graphicx}
\usepackage{subfig}
\usepackage{mathptmx} 
\usepackage{cuted}
\usepackage{lipsum, array, booktabs}
\usepackage{multirow}
\usepackage{hyperref}
\hypersetup{
    colorlinks=false,
    linkcolor=blue,
    filecolor=magenta,      
    urlcolor=cyan,
}
\usepackage{times} 
\usepackage{amssymb}
\usepackage{mathtools}

\usepackage{bm}
\usepackage{ragged2e}
\usepackage{ragged2e}

\title{\LARGE \bf
Learning to Navigate Autonomously in Outdoor Environments : MAVNet
}

\author{Saumya Kumaar$^{2}$, Arpit Sangotra$^{3}$, Sudakshin Kumar$^{3}$, Mayank Gupta$^{3}$, Navaneethkrishnan B$^{2}$ and S N Omkar$^{1}$
\thanks{$^{1}$ Chief Research Scientist, Indian Institute of Science, Bangalore}
\thanks{$^{2}$ Research Assistant, Indian Institute of Science, Bangalore}
\thanks{$^{3}$ Student, National Institute of Technology, Srinagar}
}
\begin{document}

\maketitle
\thispagestyle{empty}
\pagestyle{empty}

\begin{abstract}
In the modern era of automation and robotics, autonomous vehicles are currently the focus of academic and industrial research. With the ever increasing number of unmanned aerial vehicles getting involved in activities in the civilian and commercial domain, there is an increased need for autonomy in these systems too. Due to guidelines set by the governments regarding the operation ceiling of civil drones, road-tracking based navigation is garnering interest . In an attempt to achieve the above mentioned tasks, we propose an imitation learning based, data-driven solution to UAV autonomy for navigating through city streets by learning to fly by imitating an expert pilot. Derived from the classic image classification algorithms, our classifier has been constructed in the form of a fast 39-layered Inception model, that evaluates the presence of roads using the tomographic reconstructions of the input frames. Based on the Inception-v3 architecture, our system performs better in terms of processing complexity and accuracy than many existing models for imitation learning. The data used for training the system has been captured from the drone, by flying it in and around urban and semi-urban streets, by experts having at least 6-8 years of flying experience. Permissions were taken from required authorities who made sure that minimal risk (to pedestrians) is involved in the data collection process. With the extensive amount of drone data that we collected, we have been able to navigate successfully through roads without crashing or overshooting, with an accuracy of 98.44\%. The computational efficiency of MAVNet enables the drone to fly at high speeds of upto 6m/sec. We present the same results in this research and compare them with other state-of-the-art methods of vision and learning based navigation.
\end{abstract}

\section{Introduction}

With the advent of drone technology to the  civilian market, drones have been used in a variety of applications such as agricultural crop monitoring, surveillance, emergency first-response and delivery [1], [2], [3] and [4]. Autonomous navigation of such systems is of utmost importance to maximize mission efficiency and safety. 
 
The commonly used method of GPS waypoint-to-waypoint navigation [5] is not feasible in urban environments due to current government regulations on the altitude at which civilian drones are allowed to operate which is lower than the average height of buildings in most cities. Navigation in such cluttered environment can be implemented by simultaneous localization and mapping (SLAM) techniques [6]. However, though SLAM techniques have shown great localization prowess, issues like inertial measurement unit (IMU) sensor noise, dynamic obstacles and sharp features due to differential lighting may cause the system failure which could prove hazardous to general public. Further, advanced SLAM techniques for outdoor navigation requires specialized equipment like LIDAR and stereo-camera which are expensive and may not be compatible with standard off-the-shelf drones. 

Research on deep learning based drone navigation methods in the past decade have shown promising results. One of the interesting pieces of research by Kim \textit{et. al.} [7] proposed a deep neural network system for indoor navigation of a quadrotor drone to find a specified target. The research used monocular camera for environment perception and the images were fed into a deep convolutional neural network and the model provides the control outputs. Another method proposed by Gandhi \textit{et. al.} [8], demonstrates negative training based navigation trained on a large collection of crash dataset. This method essentially teaches the system how \textbf{NOT} to fly. It uses a concept termed as Imitation Learning.

Imitation learning is a pedagogical approach that aims to mimic human behaviour through  mapping of observations to the consecutive action performed by the human expert [9]. This method has seen numerous applications in robotics, for example, robotic arm actuation for picking up items or flipping a pancake [10]. Imitation learning has also been demonstrated in quadrotor drones for navigation through forested environment by Ross \textit{et. al.} [11]. The learning policy focuses on obstacle avoidance by modeling the human reactive control in the presence of tree-trunks. A recent piece of research by Loquercio \textit{et. al.} [12] which in line with ours, uses a convolutional neural network to navigate a drone through urban streets following basic rules like avoiding obstacles, vehicles and pedestrians. 

Unlike previously stated methods, we aim to provide a high-level real-time navigational method for a drone to travel from a specified point A to point B in an urban environment by tracking the road linking the two points using the imitation learning methodology. In other words, we teach the drone how to navigate between two points by an expert human piloting the drone between the two points. The drone estimates a mapping from an image from its monocular camera to the control command provided by an experienced human pilot.  The main contributions of this paper are:

\textbullet{}   We propose a nested convolutional network model architecture based on the Inception V3 model: Mini Aerial Vehicle Network (MAVNet) which provides five outputs namely: forward, yaw-left, yaw-right, halt and junction, and the input being a feature-extracted, reconstructed \textit{tomographic} video frame of size 100$\times$100 pixels. The junction flag is raised when a road junction is detected which enables a landmark based navigational model.

\textbullet{} The model is benchmarked on the Udacity dataset and its performance is compared with other popularly used network architectures.    

\textbullet{} The proposed architecture is also tested on a custom dataset of a 357m long path. The model was used to implement the navigational system on an off-the-shelf drone and we also show the difference in the paths followed by the expert pilot and the trained model.

The navigation system training is done with the drone's height above ground level (AGL) kept constant at 2.5m for minimal interaction with cars or pedestrians. This is done because it has been observed that the region above roads from 2m AGL to 10m AGL have fewer obstacles than when flying beyond the said range. A simple optical flow based collision avoidance system can be used in conjunction with our proposed algorithm to deal with the sparsely occurring obstacles in the said range like trucks, vans, traffic lights, tree branches etc.

\section{Methodology}
As suggested previously, in this paper, we have attempted to present an end-to-end solution for autonomous navigation through urban streets. The algorithm works in a way very similar to the learning behaviour in human beings. When the image is captured it is processed along with velocity commands given. The algorithm learns a controller that maps the input key-strokes to the image frame. Once the processing is done, the MAVNet model predicts a total of 5 values in the band. The last value of the array is an indicator of presence or absence of a junction. So, whenever the array is 1, a counter keeps a track of how many junctions have passed by counting the number of 1's. When the count reaches the required value, the drone is given a suitable yaw command (either left or right, which is not a MAVNet prediction but pre-programmed by the user) which allows it to make appropriate turns at the junctions. This landmark based method is how human brain processes navigational information. We attempt to implement the same in unmanned aerial vehicles, a process flow of which is represented graphically in Fig. 1

\begin{figure}[htp]
\centering
\includegraphics[width=3.3 in]{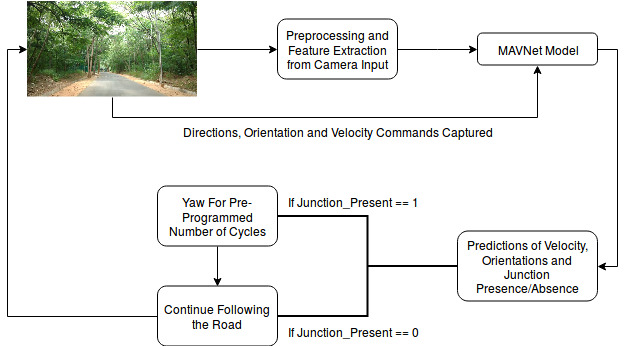}
\caption{Overall Architecture of the MAVNet Prediction System.}
\label{fig_chi_dot}
\end{figure}

In contrast to \textit{Loquercio et. al.} [12], the collision prediction tasks are not tackled in this research as the height of the UAV was fixed at 2.5m. At this height, most of the on-track vehicles and pedestrians do not interact with the drone, and any observed poles or tree branches were above the fixed height in the collected custom dataset, so the need for collision avoidance was not critical. So the MAVNet model predicts only the forward pitch rate, the yaw angle and the probability of the presence of a road junction in the current frame. Apart from the above mentioned outputs, we have one more additional output, which we call \textit{halt} which means that the UAV has to hover in its place without any movement. So although the training was not done for avoiding obstacles, but the training set included \textit{no commands} being given to a drone, whenever there was an obstacle in the surrounding. This kind of imparts an inherent collision avoiding capabilities to the drone, however we don't report or evaluate it.

\subsection{Simulation Environments}
Prior to the implementation of our algorithm onto a real-world interface, we created a simulation environment using commercial desktop entertainment programs like GTA San Andreas(GTA SA) and Mario Kart. This was done because deploying the algorithm directly on a UAV without any simulation testing, could pose a serious risk to passers-by, vehicles or other property. However, we did not perform any feature extraction experiments with the simulation data. We used unprocessed raw images captured from recording the screen and the key-strokes. In GTA SA we collected our data on an oval shaped circuit following a white demarcation line on the road. We trained our model on 48 video segments each containing approximately 10,000 frames, whereas for Mario Kart We collected 120 video segments each containing approximately 3,400 frames. In both the cases, the model learned to play quite similar to the human player. With the confidence acquired from the simulations, we decided to recreate the same with real-world applications. The model simply learns how the expert flies the drone by learning the key-strokes rather than any other complex features.

\begin{figure}
    \centering
    \subfloat{{\includegraphics[width=2cm]{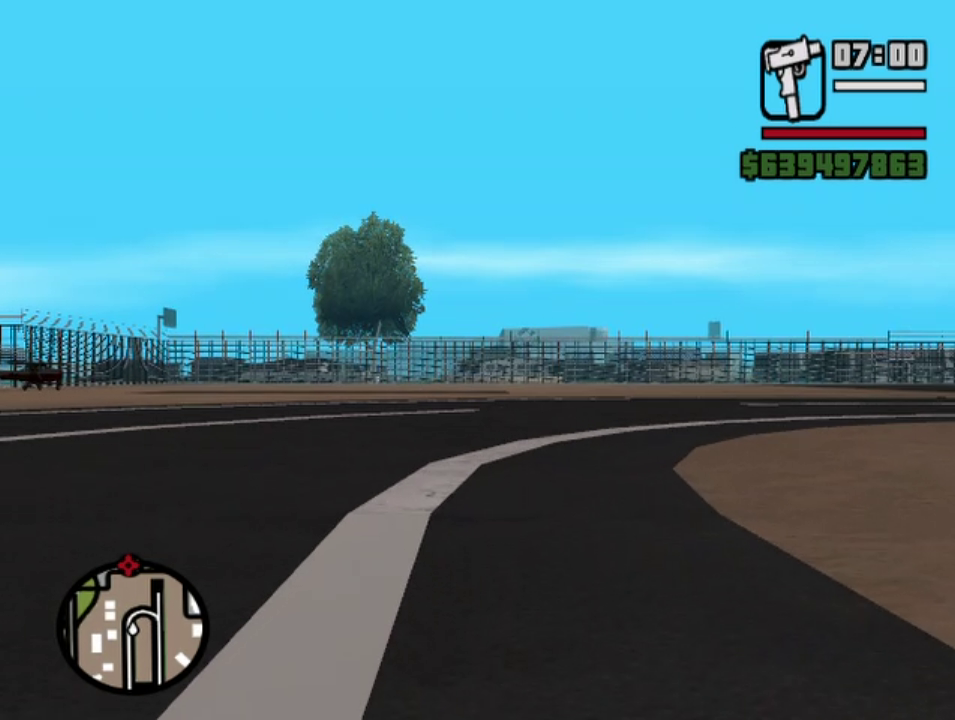} }}
    \subfloat{{\includegraphics[width=2cm]{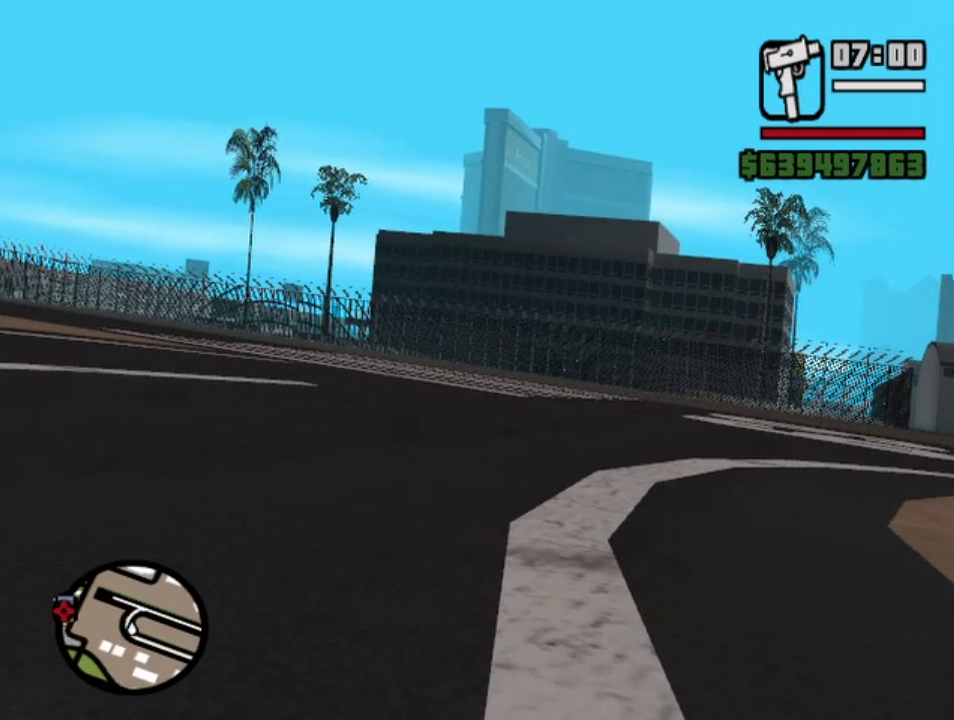} }}
    \subfloat{{\includegraphics[width=2cm]{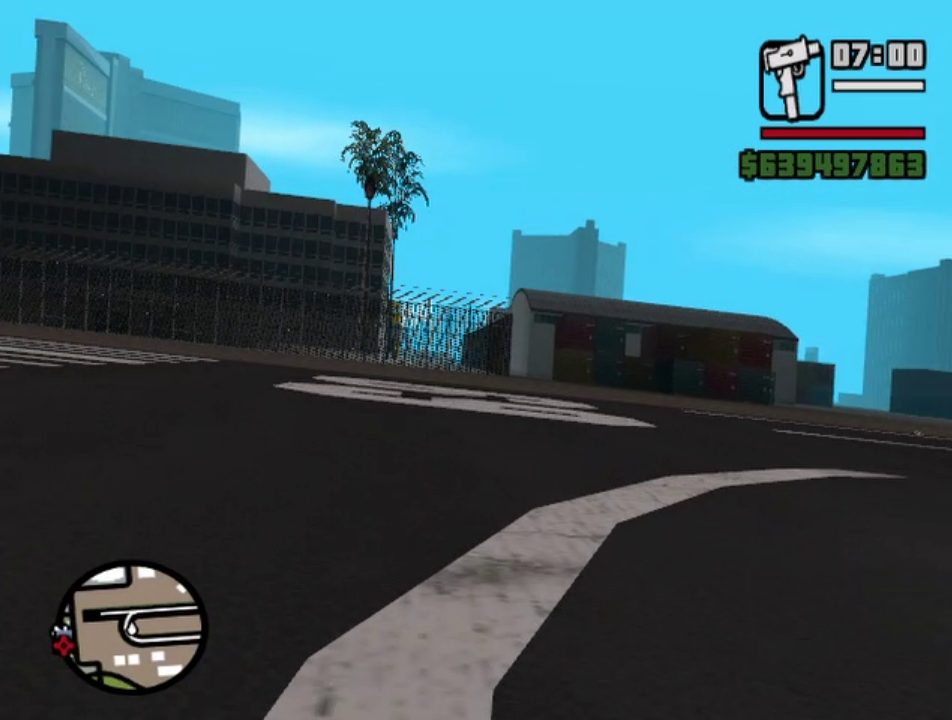} }}
    \subfloat{{\includegraphics[width=2cm]{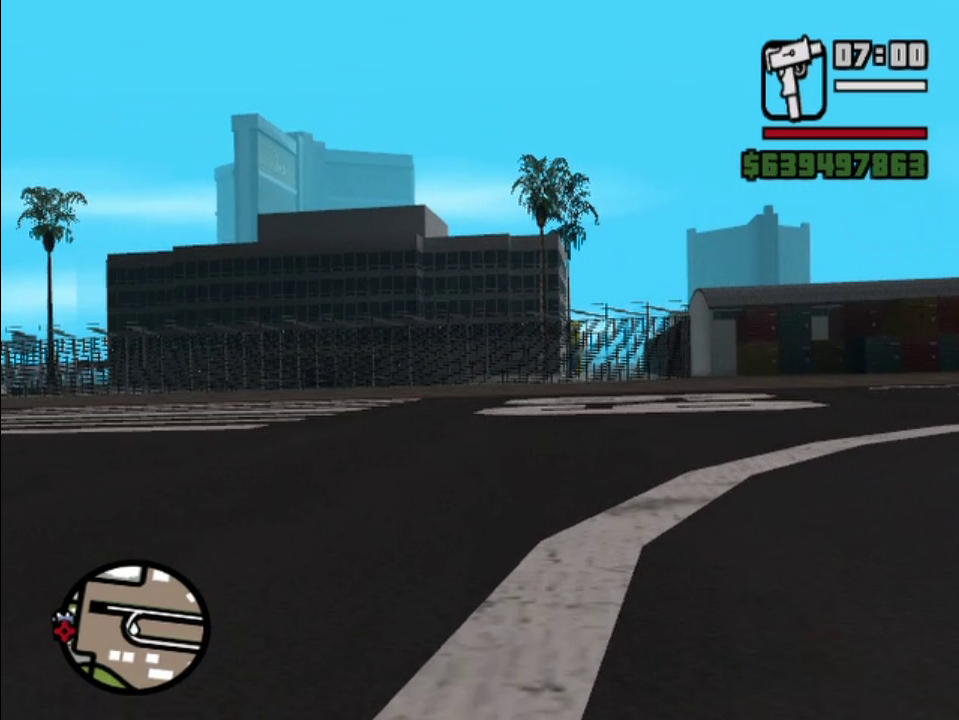} }}
        \quad
    \subfloat{{\includegraphics[width=2cm]{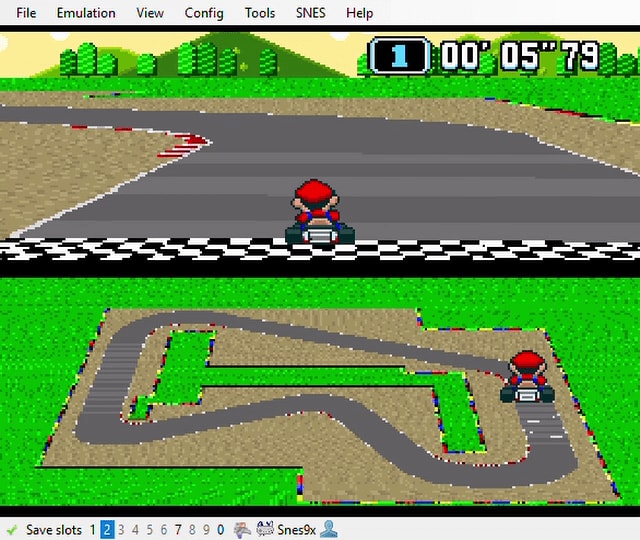} }}
    \subfloat{{\includegraphics[width=2cm]{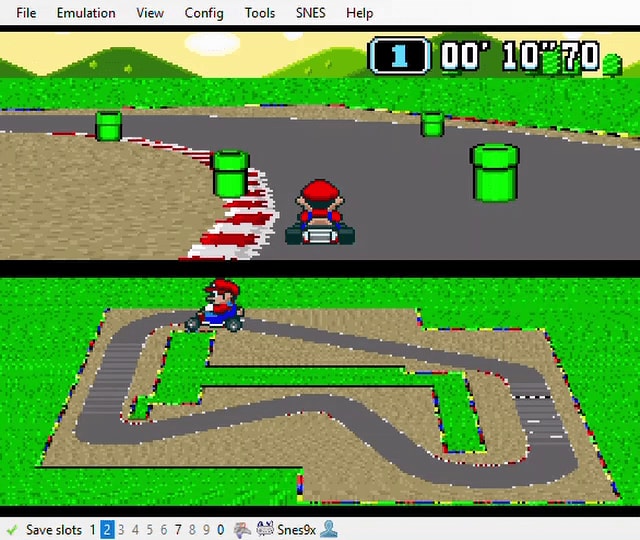} }}
    \subfloat{{\includegraphics[width=2cm]{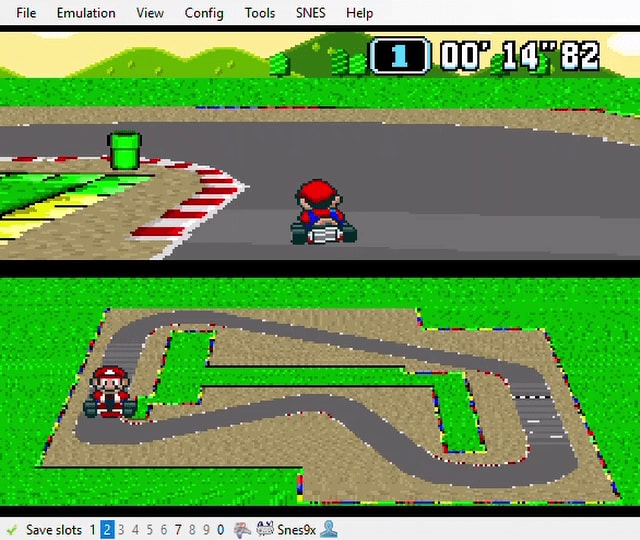} }}
    \subfloat{{\includegraphics[width=2cm]{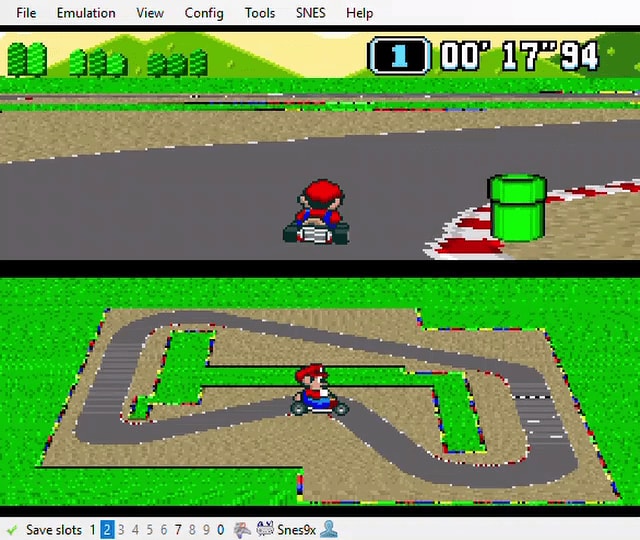} }}
    \caption{Sample Testing Instances from both the simulation environments. On the top, the figures indicate the GTA San Andreas environment where our system over-fitted on the \textit{forward} command. The lower portion of the figure indicates the Mario Kart environment which performed way better than previous one primarily because of more features available in the image dataset. In both the cases, we collected around 400,000 data frames for training, with the logic that more the training, better the performance would be.}
    \label{fig:example}
\end{figure}

\subsection{Datasets and Preprocessing}
There are two datasets involved in this research. One is the Udacity's Self-Driving simulation environment. A part of the dataset (which contains around 72,000 images) has been used for training and testing is done on unseen roads of the test dataset. The learnt parameters were throttle and steering angles. The key-strokes used for turning the car in the Udacity's self-driving simulator were deemed as the corresponding yaw-commands for the drone. This further helps in classification as there are only two classes are possible for steering. However, due the absence of road junctions in the Udacity dataset, we establish the metrics on the Cityscapes [19] and our own dataset, where every image has associated velocity inputs and junction tags. The custom database is composed of around 450,000 images, in three different times of the day, with varying sunlight, shadows and traffic conditions. There were a total of 71 flights conducted during the course of the research.

\begin{figure}
    \centering
    \subfloat{{\includegraphics[width=2.5cm]{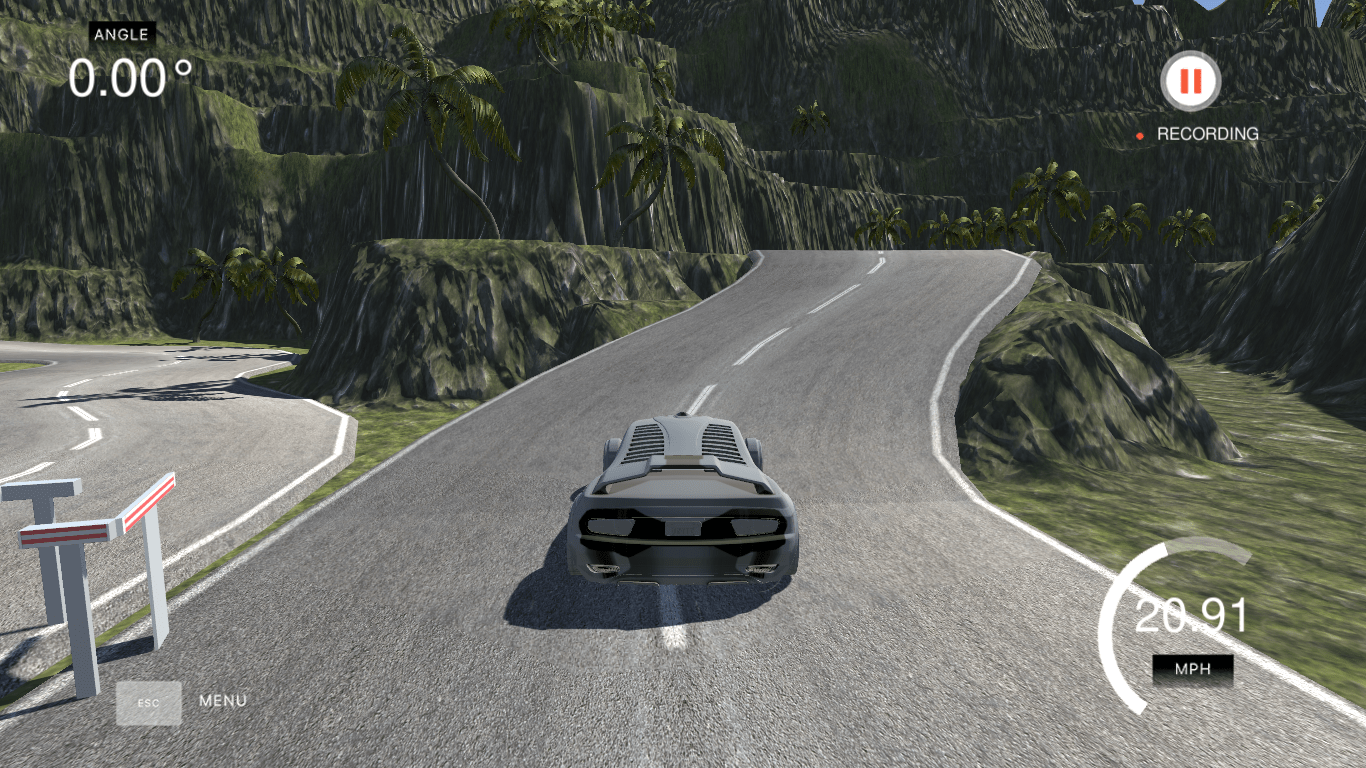} }}
    \subfloat{{\includegraphics[width=2.5cm]{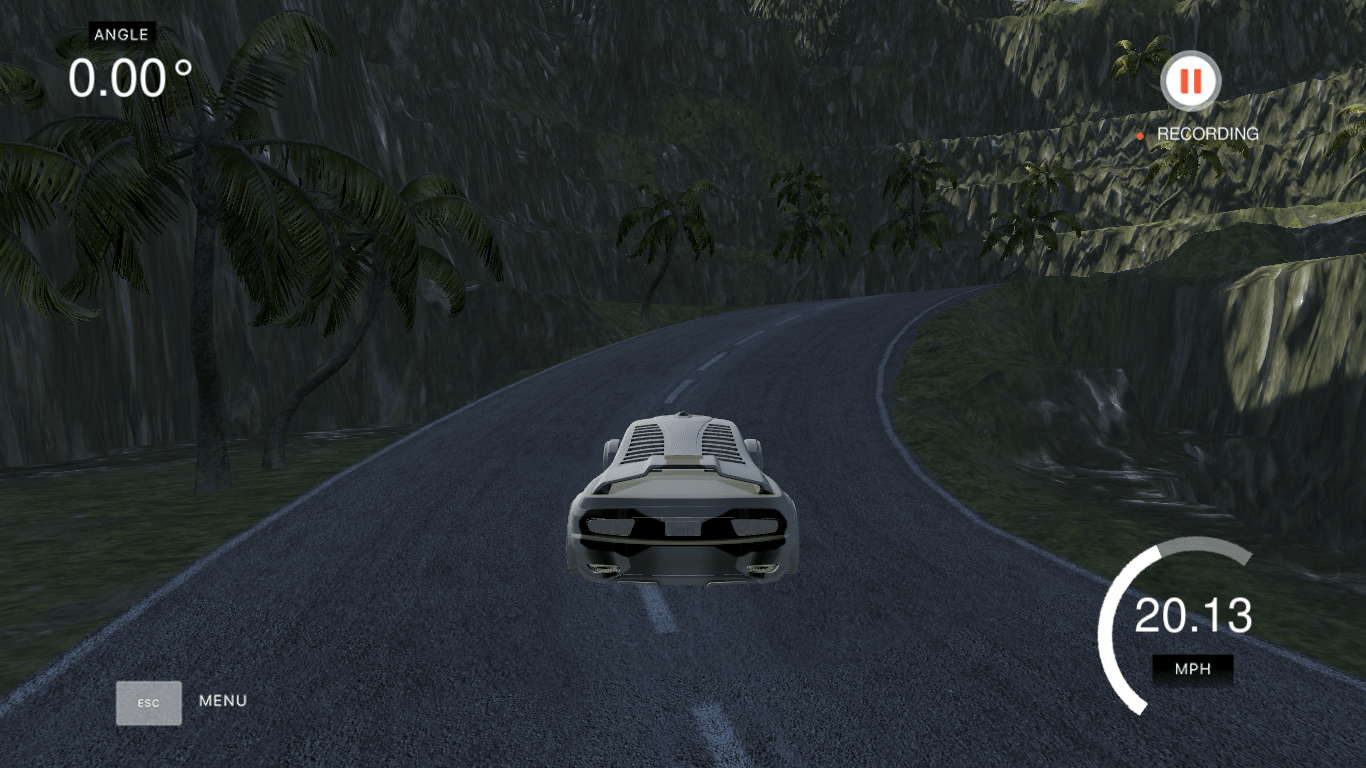} }}
    \subfloat{{\includegraphics[width=2.5cm]{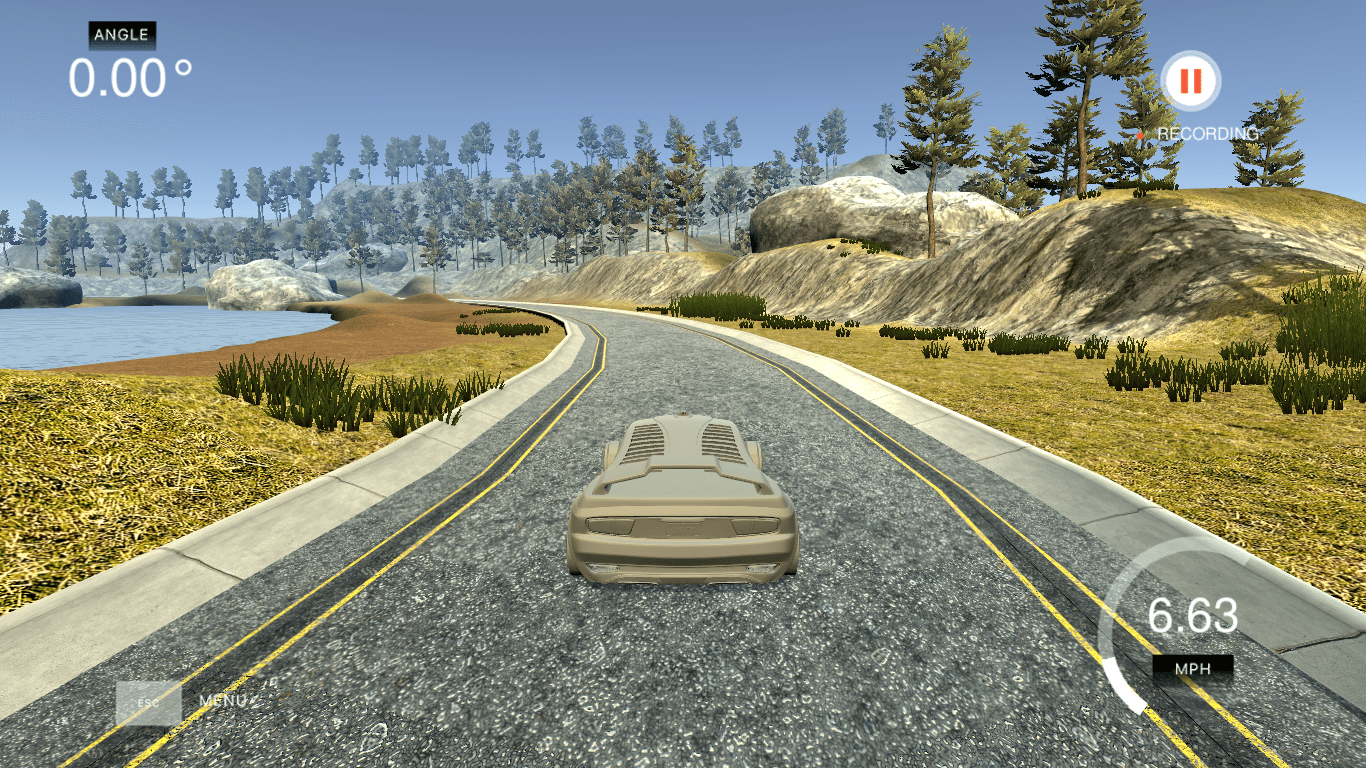} }}
        \quad
    \subfloat{{\includegraphics[width=2.5cm]{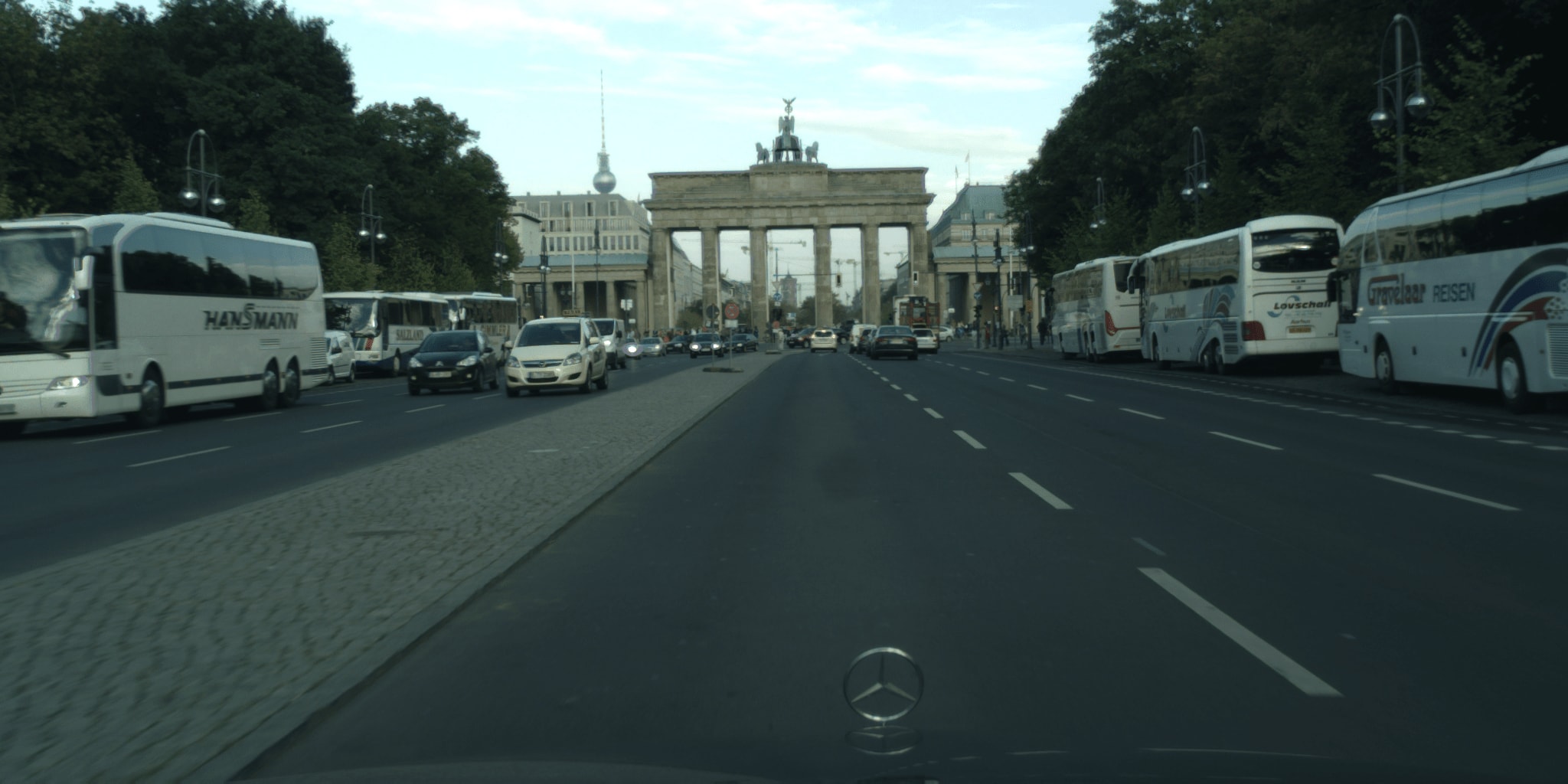} }}
    \subfloat{{\includegraphics[width=2.5cm]{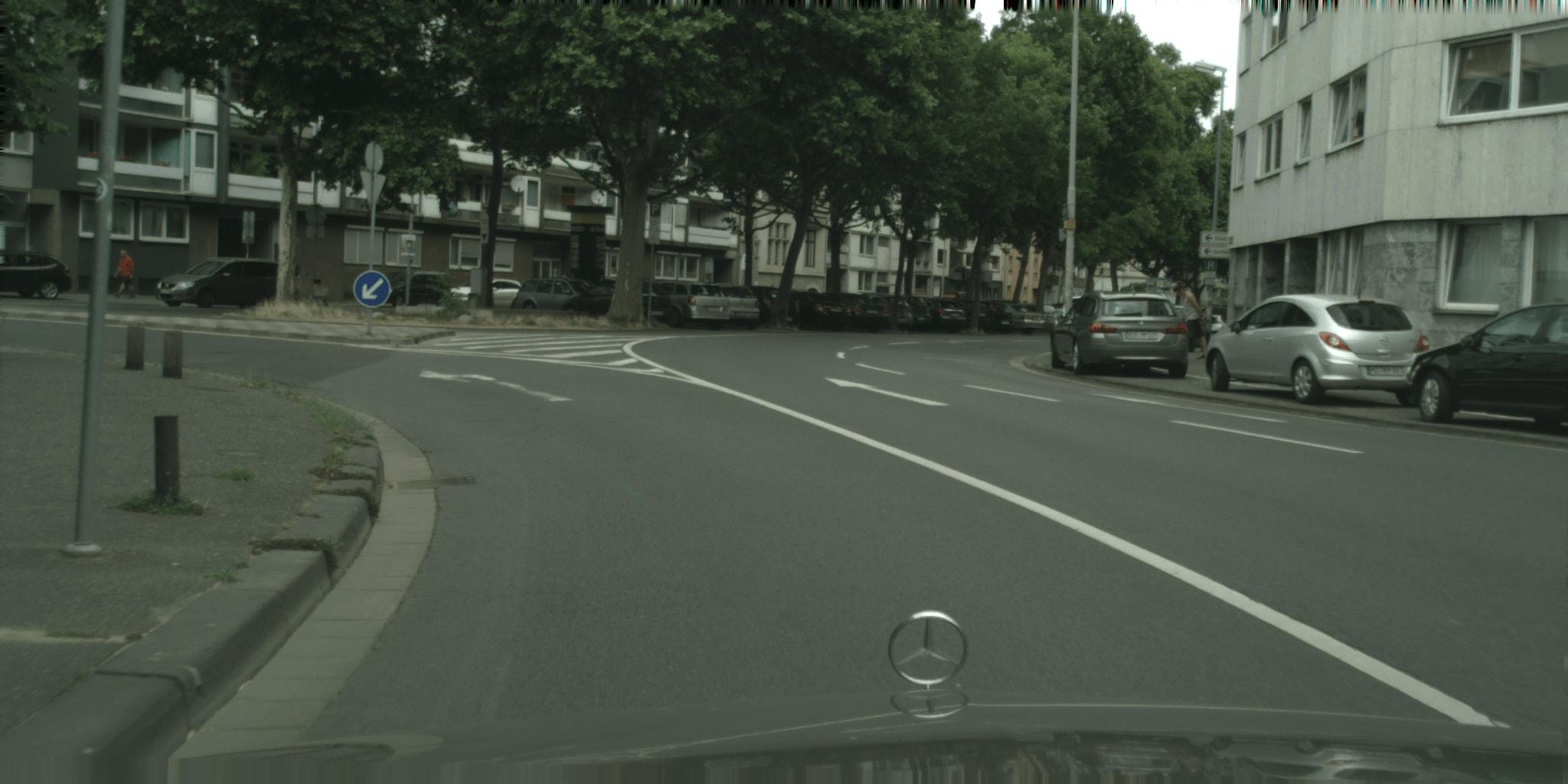} }}
    \subfloat{{\includegraphics[width=2.5cm]{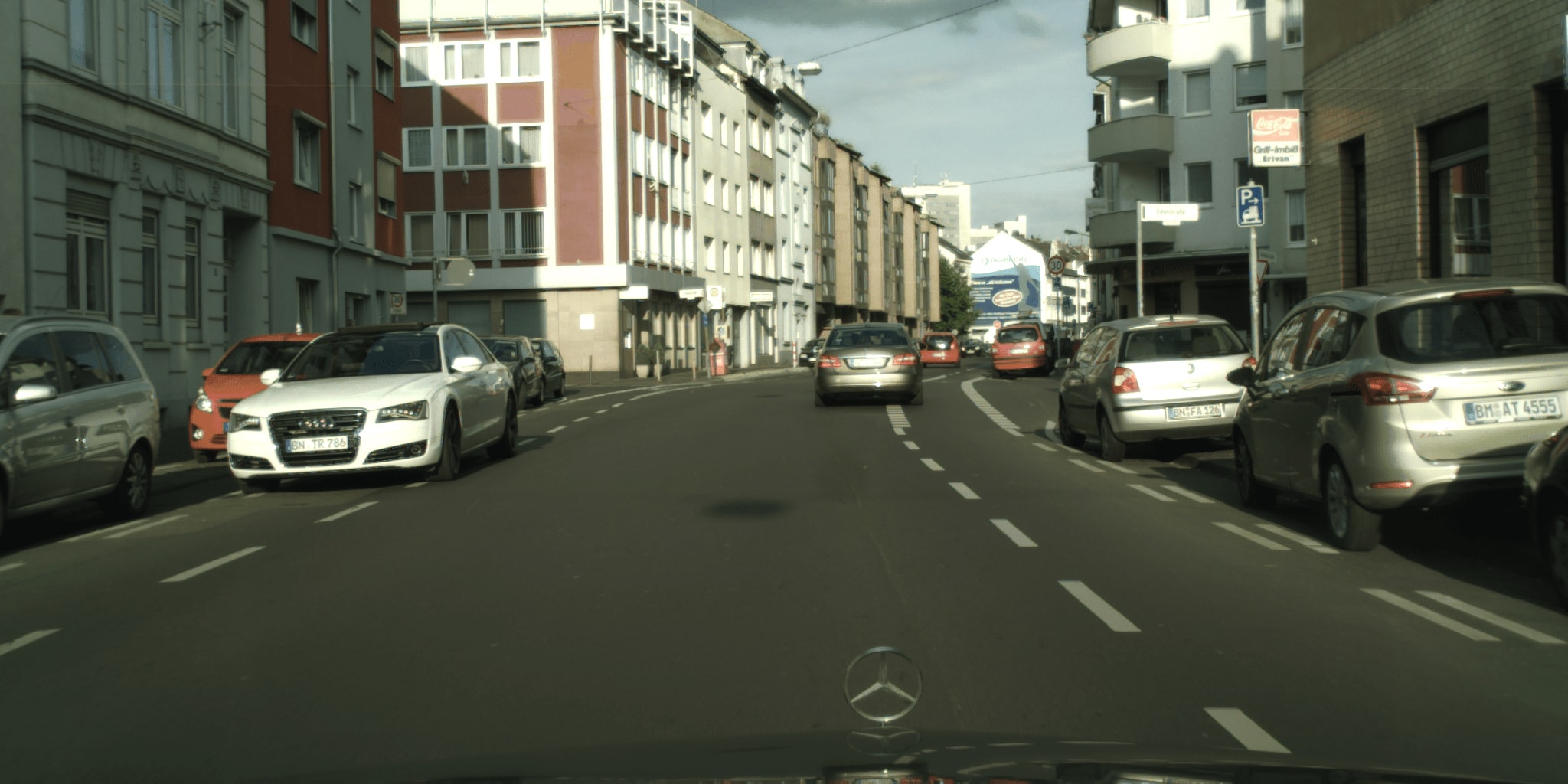} }}

    \caption{Training Images as collected from the Udacity Dataset and the Cityscapes Dataset. In order to benchmark our algorithm against some of the state-of-the-art techniques, our MAVNet model learns the throttle and steering (as binarized left/right) commands from the Udacity Dataset and junction detection tasks from the Cityscapes dataset. Evaluation has been performed on both datasets with separate metrics for both.}
    \label{fig:example}
\end{figure}

\begin{figure}
    \centering
    \subfloat{{\includegraphics[width=2cm]{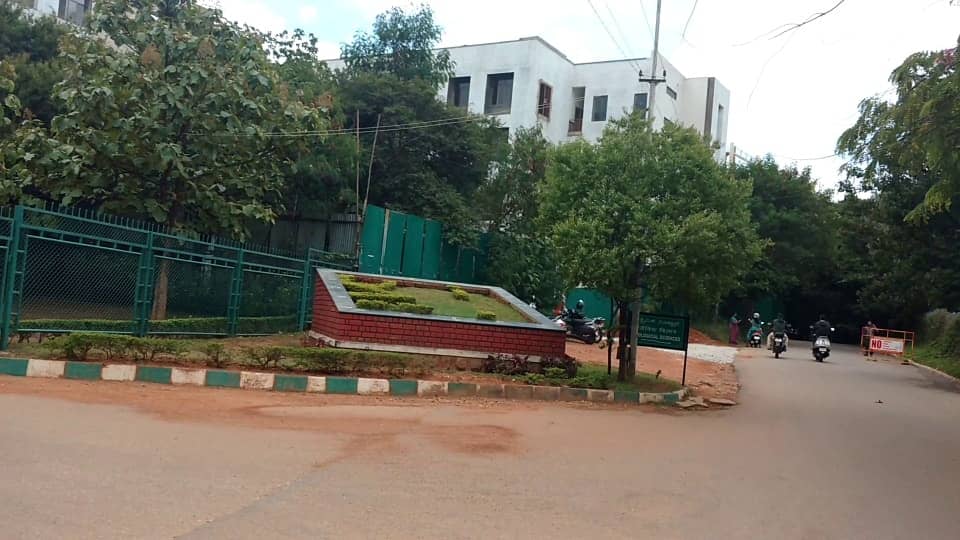} }}
    \subfloat{{\includegraphics[width=2cm]{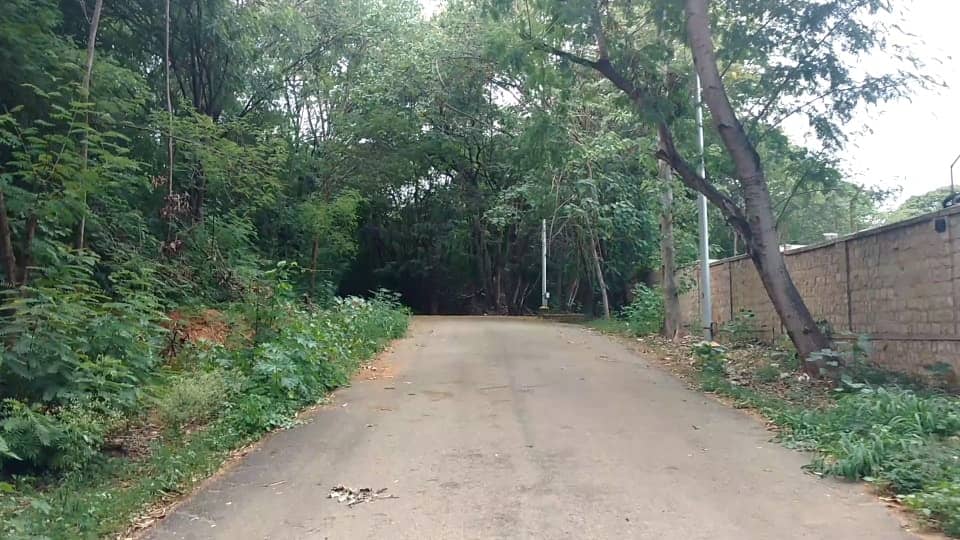} }}
    \subfloat{{\includegraphics[width=2cm]{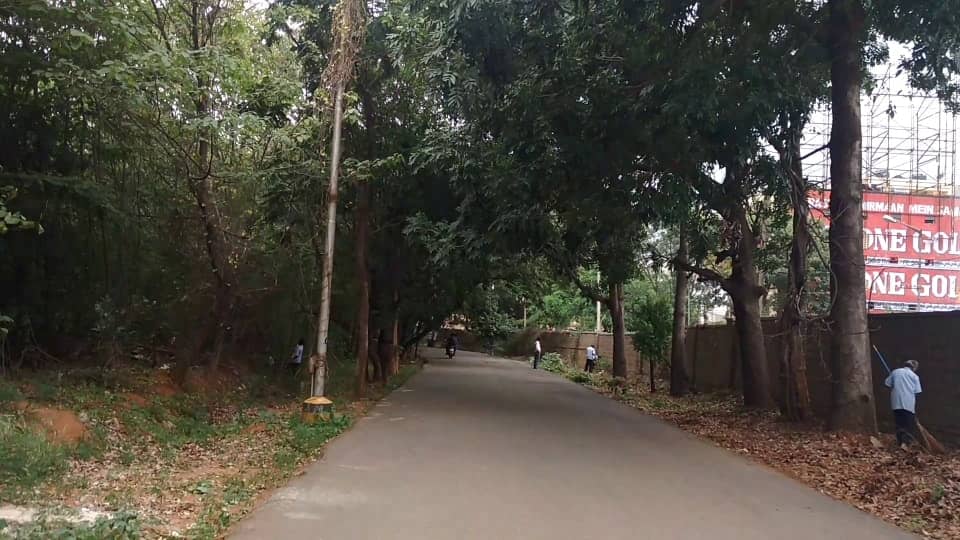} }}
    \subfloat{{\includegraphics[width=2cm]{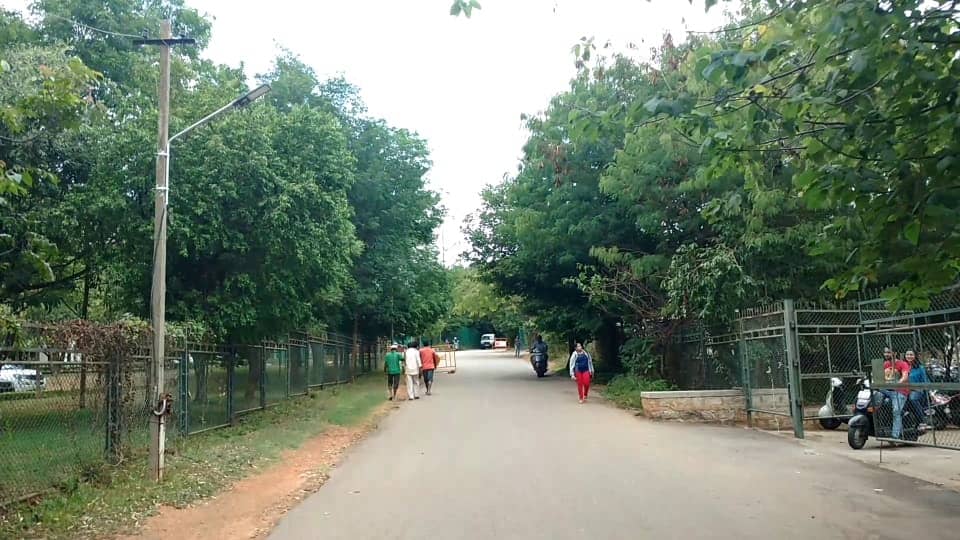} }}
        \quad
    \subfloat{{\includegraphics[width=2cm]{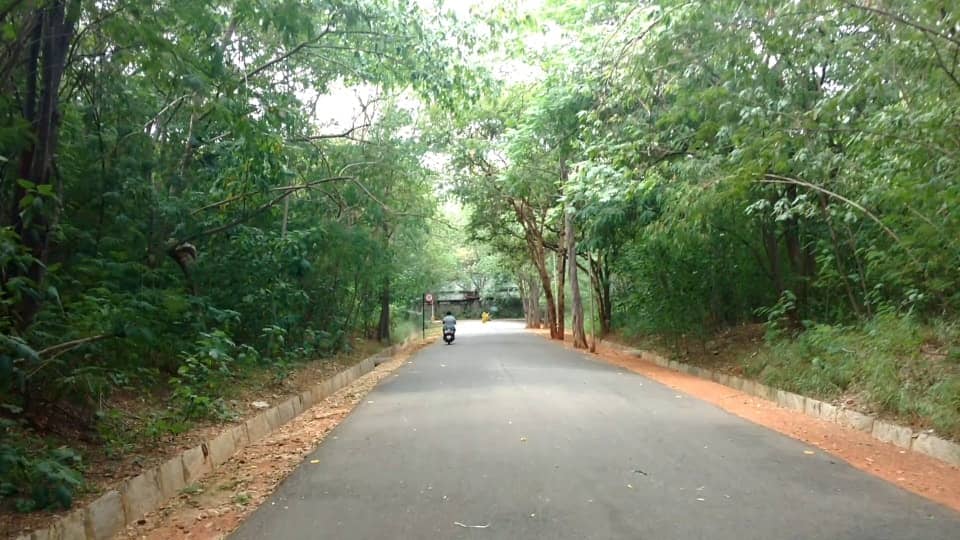} }}
    \subfloat{{\includegraphics[width=2cm]{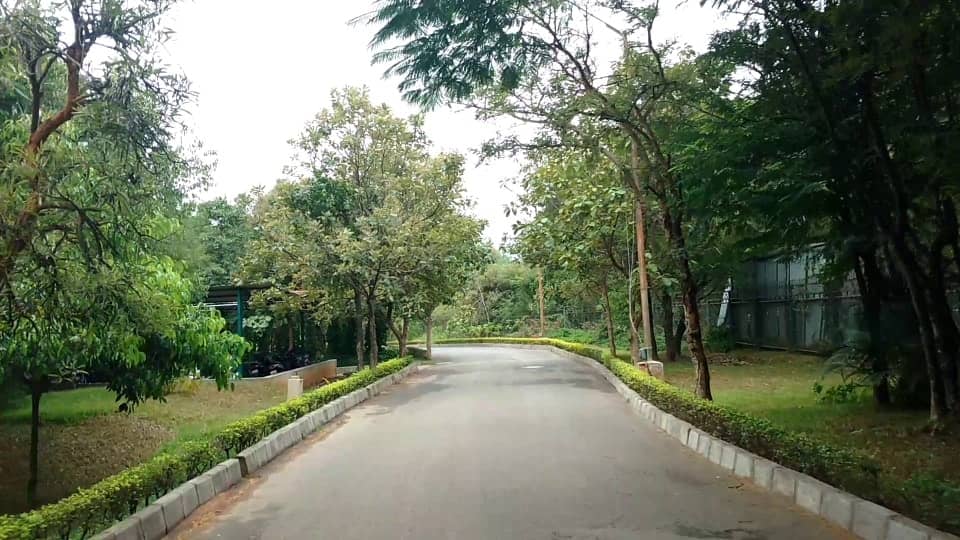} }}
    \subfloat{{\includegraphics[width=2cm]{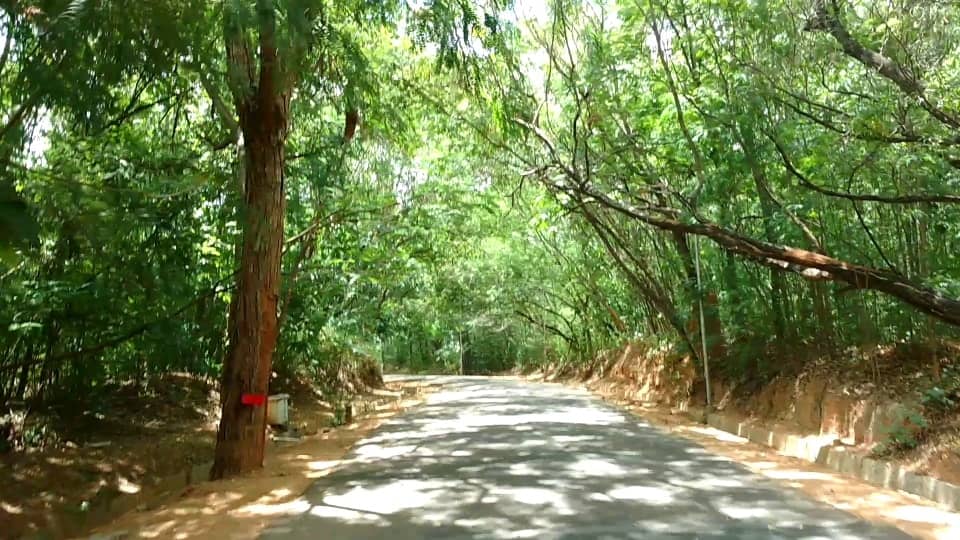} }}
    \subfloat{{\includegraphics[width=2cm]{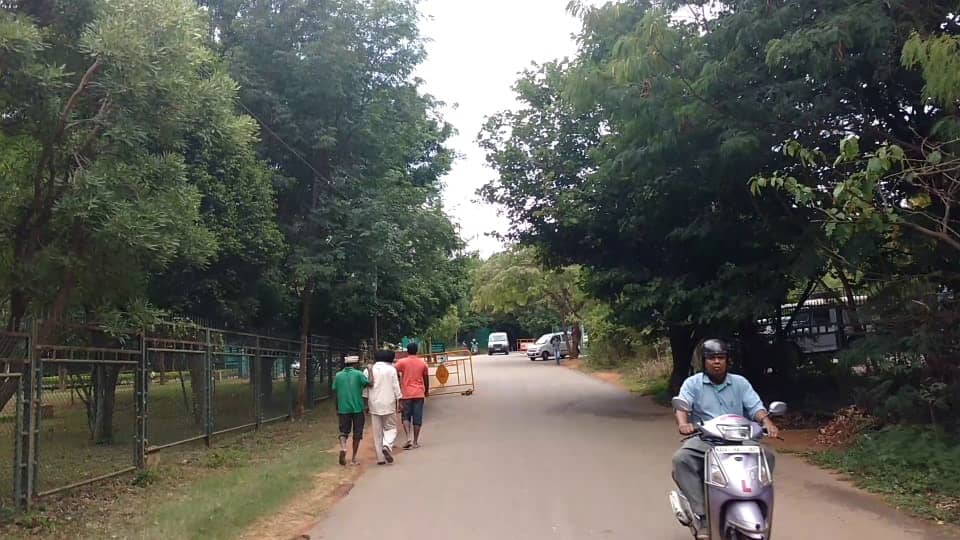} }}
            \quad
    \subfloat{{\includegraphics[width=2cm]{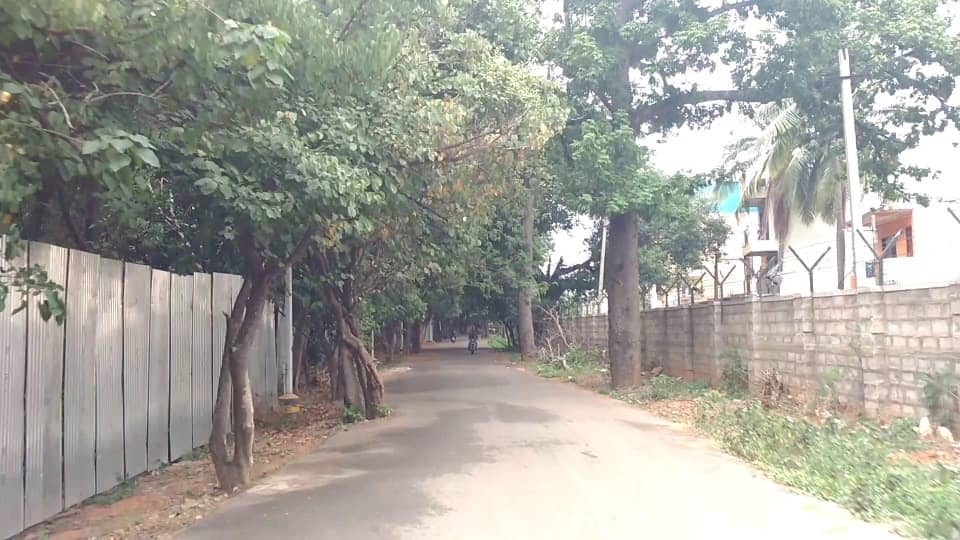} }}
    \subfloat{{\includegraphics[width=2cm]{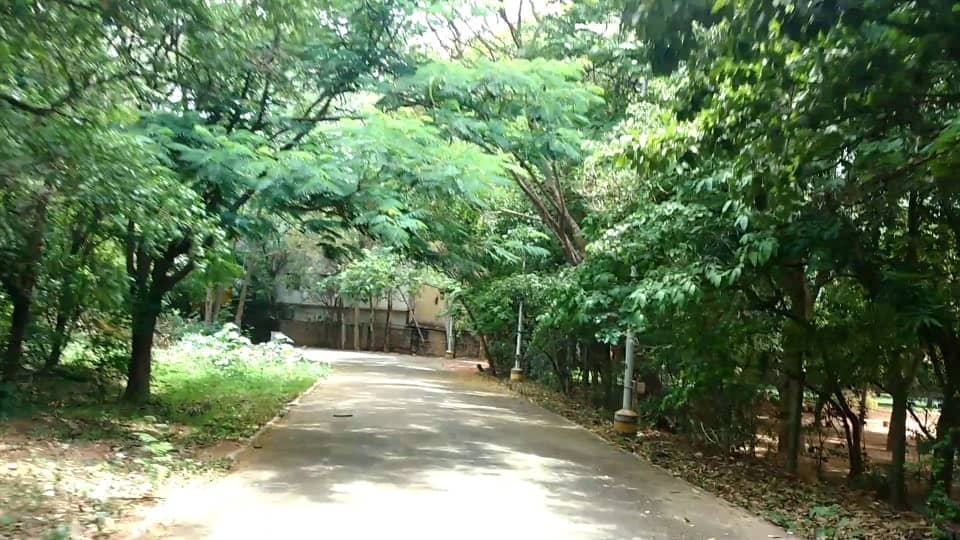} }}
    \subfloat{{\includegraphics[width=2cm]{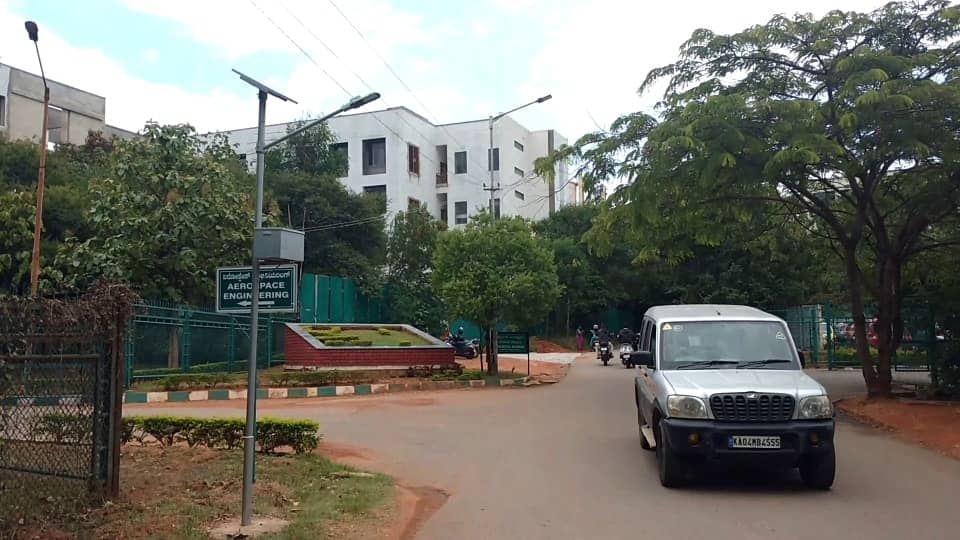} }}
    \subfloat{{\includegraphics[width=2cm]{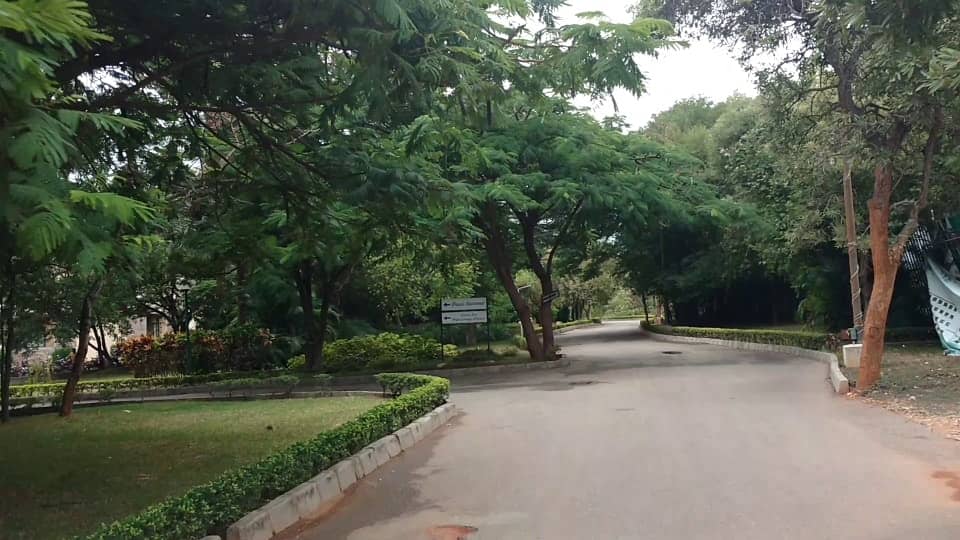} }}
    \caption{Training Images as collected in and around the campus of Indian Institute of Science. The dataset contains a variety of junctions and marked, unmarked, structured and unstructured roads. The training is done extensively on these and test results are presented exclusively on untrained roads. The dataset was collected during different times of the day, under different sunlight conditions.}
    \label{fig:example}
\end{figure}

\subsection{Feature Extraction}

In this research, we have proposed to extract the  \textit{radon features} of the visual data. Although the sinogram so constructed could be used effectively for road segmentation, we take a step ahead and reconstruct the image which helps in complete isolation of the road from the rest of the environment. Fundamentally, roads have the least amount of edge energies when compared to its surroundings. The Radon transformation technique has proven to be effective in curved segment detection in images and reconstruction of tomographic images by Toft \textit{et. al.} [14]. The approach adopted for the above mentioned tasks were stated by Deans \textit{et. al.} [13], and the same has been described further in this section : 

\subsubsection{Radon Transformation} 

Let us say that $f(\textbf{x}) = f(x,y)$ is a continuous function that is compactly supported on \textbf{R}$^2$. The Radon transform, $Rƒ$, is therefore a function defined on the total or overall space of straight lines $L$ in \textbf{R}$^2$ by the line integral along every such line in the space:

\begin{equation}
Rf(L) = \int_L f(x) |dx|
\end{equation}

Significantly, the coefficients (parameters) of any straight line $L$ with respect to length of the arc $z$ can always be shown to be:

\begin{equation}
(x(z),y(z)) = \big((z\sin\alpha + s\cos\alpha),(-z\cos\alpha + s\sin\alpha)\big)
\end{equation}

where $s$ is the distance of this line $L$ from the origin $O$ and $\alpha$ is the angle between the normal vector to $L$ and the $x$ axis. It follows that the angular and distance physical quantities such as ($\alpha,s$) can be considered or deemed as spatial coordinates on the space of every such line in \textbf{R}$^2$, and the Radon transform can be calculated (just like the Fast Fourier Transform) in these coordinates by the following methods:

    \begin{align*}
Rf(\alpha, s) = & \int_{-\infty}^{\infty} f(x(z),y(z)) dz \\
= & \int_{-\infty}^{\infty} f((z\sin\alpha + s\cos\alpha),(-z\cos\alpha + s\sin\alpha)) dz
    \end{align*}

Speaking generally, in the overall n-dimensional Euclidean space \textbf{R}$^n$, the Radon transformation of a compactly supported continuous function mentioned as $f$ is a function $Rf$ on the space $\Sigma_n$ of all hyperplanes in \textbf{R}$^n$. It is defined as :

\begin{equation}
Rf(\xi) = \int_\xi f(x) |dx|
\end{equation}

for $\xi$ $\in$ $\Sigma_n$, where the integral of the function is calculated with respect to the natural hyper-surface estimate, $d\sigma$ (generalizing the $|d\textbf{x}|$ term from the 2-dimensional case). It is definitely worth noting that any element of $\Sigma_n$ can be classified as the solution locus of the following equation :

\begin{equation}
\textbf{x}.\alpha = s
\end{equation}

where $\alpha$ $\in$ $S^{n-1}$ is a unit vector and $s \in R$. Thus the n-dimensional Radon transformation matrix may be rewritten as a function on $S^{n-1}\times\textbf{R}$ via

\begin{equation}
Rf(\alpha,s) = \int_{x.\alpha=s}f(x)d\sigma(\textbf{x}).
\end{equation}

We can also consider a generalized Radon transform still further by integrating instead over the k-dimensional affine subspaces of $R^n$. The X-ray transform is the most commonly used special case of this construction, and is obtained by integrating over straight lines, again a point worth noting. 

\subsubsection{Reconstruction Approach - Ill-posedness}

As mentioned previously, we reconstruct the image for better accuracy of classification. The process of reconstruction that has been adopted here is done using \textit{Ill-posedness} which produces the image (or function $f$ in the previous section) from its projection data. Reconstruction is fundamentally an inverse problem. The above mentioned approach is used because of its computationally effectiveness for the Radon transform. The Radon transform in n-dimensions can be inverted (or reconstructed finally) by the following :

\begin{equation}
c_{nf} = (-\Delta)^{(n-1)/2}R*Rf
\end{equation}

where,

\begin{equation}
c_{n} = (4\pi)^{(n-1)/2}\frac{\Gamma(n/2)}{\Gamma(1/2)}
\end{equation}

and the power of the Laplacian $-\Delta^{(n-1)/2}$ is defined as a pseudodifferential operator if necessary by the Fourier transform :

\begin{equation}
F\big[-\Delta^{(n-1)/2}\phi\big] (\xi) = |2\pi\xi|^{n-1}F\phi(\xi)
\end{equation}

For computational speed and efficiency, the power of the Laplacian is commuted with the dual transform $R^*$ to give :

\[
    c_nf= 
\begin{cases}
    R^*\frac{d^{n-1}}{ds^{n-1}}Rf,& \text{$n$ odd}\\
    R^*H_s\frac{d^{n-1}}{ds^{n-1}}Rf,              & \text{$n$ even}
\end{cases}
\]

where $H_s$ is the Hilbert transform with respect to the $s$ variable. In 2 dimensions, the operator $H_sd/ds$ is a fundamental \textit{ramp filter} in image processing techniques. We can hence, prove directly from the Fourier slice theorem and a slight change of variables for integration, that for a compactly continuous supported function $ƒ$ of 2 variables the following holds true :

\begin{equation}
f = \frac{1}{2}R*H_s\frac{\textit{d}}{\textit{ds}}Rf
\end{equation}

Therefore, in an image processing case, the original image $ƒ$ can be regenerated from the \textit{sinogram} data $Rƒ$ by applying a basic ramp filter (in the $s$ variable) and then back-projecting as discussed. As the filtering step can be performed very efficiently and effectively (for example using digital signal processing techniques and tricks) and the back projection step is simply an assimilation of values in the individual pixels of the image, this results in a highly computationally efficient, and hence widely used, algorithm.

Explicitly, the inversion formula obtained by the latter method is :

\begin{equation}
f(x) = \frac{1}{2}(2\pi)^{1-n}(-1)^{(n-1)/2}\int_{S^{n-1}}\frac{\partial^{n-1}}{\partial s^{n-1}}Rf(\alpha,\alpha\cdot x)d\alpha
\end{equation}
if $n$ is odd, and

\begin{equation}
f(x) = \frac{1}{2}(2\pi)^{-n}(-1)^{n/2}\int_{-\infty}^{\infty}\frac{1}{q}\int_{S^{n-1}}\frac{\partial^{n-1}}{\partial s^{n-1}}Rf(\alpha,\alpha\cdot x + q)d\alpha dq
\end{equation}
if $n$ is even.

\begin{figure*}
  \includegraphics[width=\textwidth]{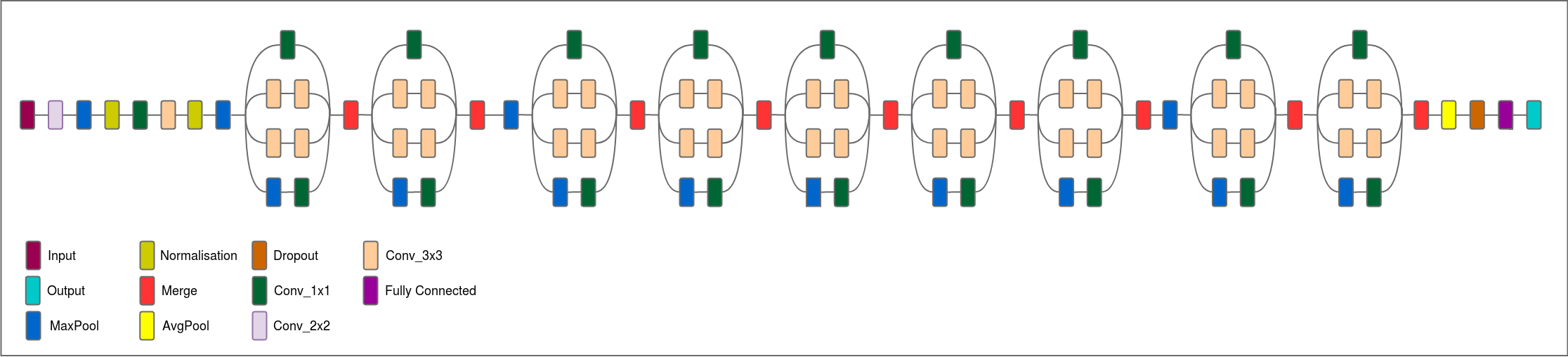}
  \caption{Our MAVNet Architecture. The multiple 1$\times$1 convolutions make sure all local features are extracted from the tomographic reconstructions. The final fully connected layer gives the final output in the form of an array that looks like :- [1, 0, 1, 0, 0]}
\end{figure*}

So the the generic Radon transform-and-reconstruction when applied to our road dataset is observed as follows. As is seen clearly, the road-junctions appear as extensive \textit{black} regions in the images whereas the other side-paths appear as \textit{white}. The tomographic reconstructed image (as shown in Fig. 4) are then fed to the neural network for classification.

\begin{figure}
    \centering
    \subfloat{{\includegraphics[width=2.1cm]{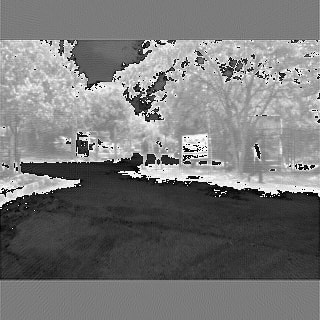} }}
    \subfloat{{\includegraphics[width=2.1cm]{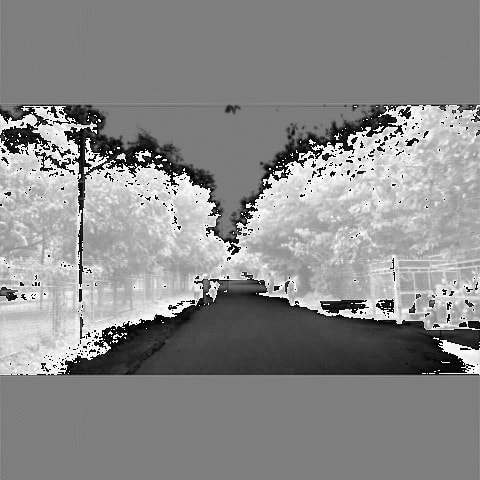} }}
    \subfloat{{\includegraphics[width=2.1cm]{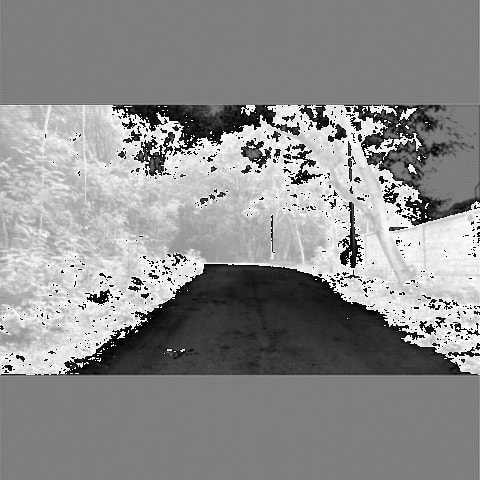} }}
    \subfloat{{\includegraphics[width=2.1cm]{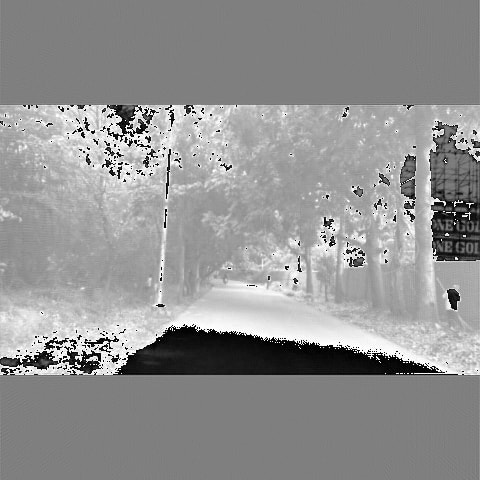} }}
        \quad
    \subfloat{{\includegraphics[width=2.1cm]{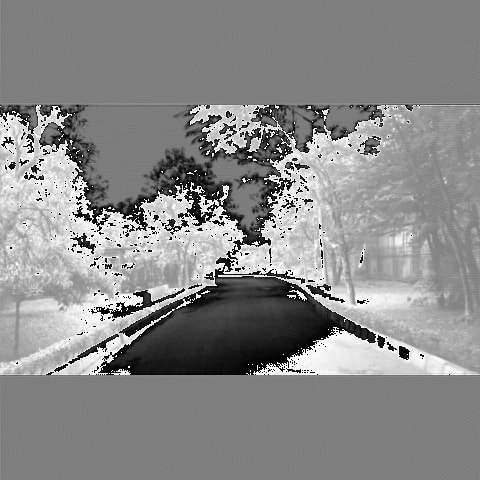} }}
    \subfloat{{\includegraphics[width=2.1cm]{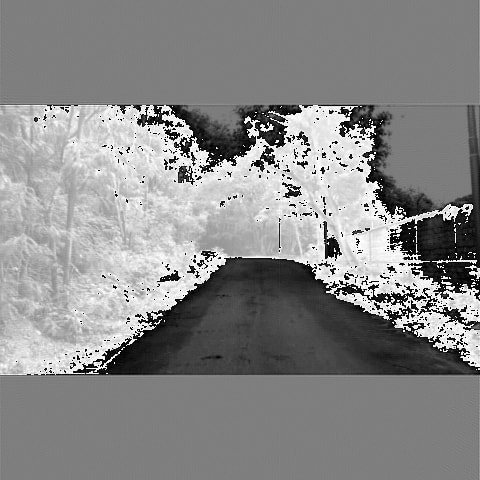} }}
    \subfloat{{\includegraphics[width=2.1cm]{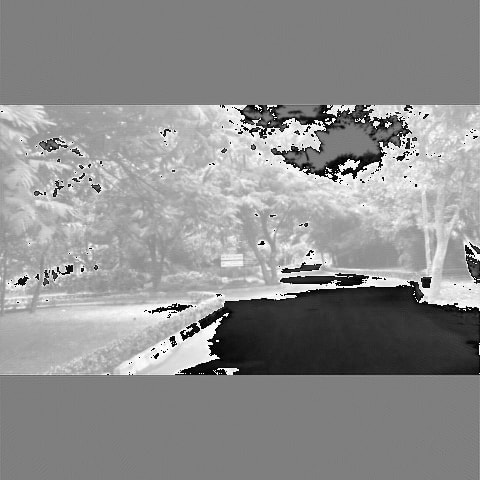} }}
    \subfloat{{\includegraphics[width=2.1cm]{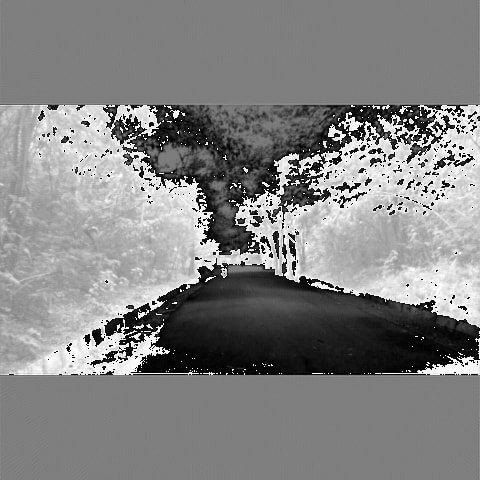} }}
    \caption{The figure shows the tomographic reconstruction of the images after applying the Radon Transform. Interestingly, the top-left image represents a road-junction. As it is clearly seen, road-junctions could be classified based on the amount of road present in the frame. A widespread distribution of road pixels (minimal edge energies) would usually indicate the presence of a junction. The rest of the images indicate roads that the drone learns to follow. These images support our claim of using X-Ray vision for navigation.}
    \label{fig:example}
\end{figure}

\subsection{The MAVNet Model}

Originally, the Inception V3 model was tried for indoor navigation via imitation learning by Szegedy \textit{et. al.} [17]. However, the research was mostly focused on simulation of the setup rather than an actual implementation in real-time. Based on this approach, we decided to try the same model for outdoor navigation. Interestingly it was observed, inception v3 performs remarkably on the custom dataset.

But as it is commonly understood, convolutional neural networks work on a basic assumption that most of the low level features of the image are local in nature, and that whatever function is applicable to one region would be applicable to others as well. So before discussing about the changes we made to the existing architecture of the inception-v3 model, let us discuss about the impact of the filter size on convolutional nets. Size of the filters plays an important role here. A larger sized filter could probably miss out on the low-level features in the images and end up skipping some important details, whereas a smaller filter could prevent that, but results in more confusion due to increased information. Now the inception model has already been benchmarked as a computationally fast classification architecture and has been proven to outperform many standard classification techniques. 

In our experiments with filter sizes in the inception-v3 model, we observed that there is a multitude of 1$\times$1 convolutional filters. However, the inception layers 3, 4 and 5 have 5$\times$5 filters also. The presence of 1$\times$1 convolutional filters reduces the need of any other filter size because every single pixel in the image contributes to the feature vector directly. So in order to reduce complexity, the convolutional layers with 5$\times$5 were removed from the architecture. Experimentally, this results in a faster computation time for the same dataset on a system with minimal hardware availability. A graphical visualization of MAVNet is shown in Fig. 9.

\section{Hardware Implementation}
As a proof of our real-time implementation we performed our real-time experiments on \textit{Parrot Bebop 2}, an off-the-shelf quadrotor commonly available in the market. The MAVNet prediction model sends only high level velocity and yaw commands to the machine for execution. Since the Bebop 2 does not support high-level on-board computing, the MAVNet model runs on a portable system with a i-3 Core processor at 2.1 GHz, without any GPU support. 

\begin{figure}
    \centering
    \subfloat{{\includegraphics[width=2.65cm]{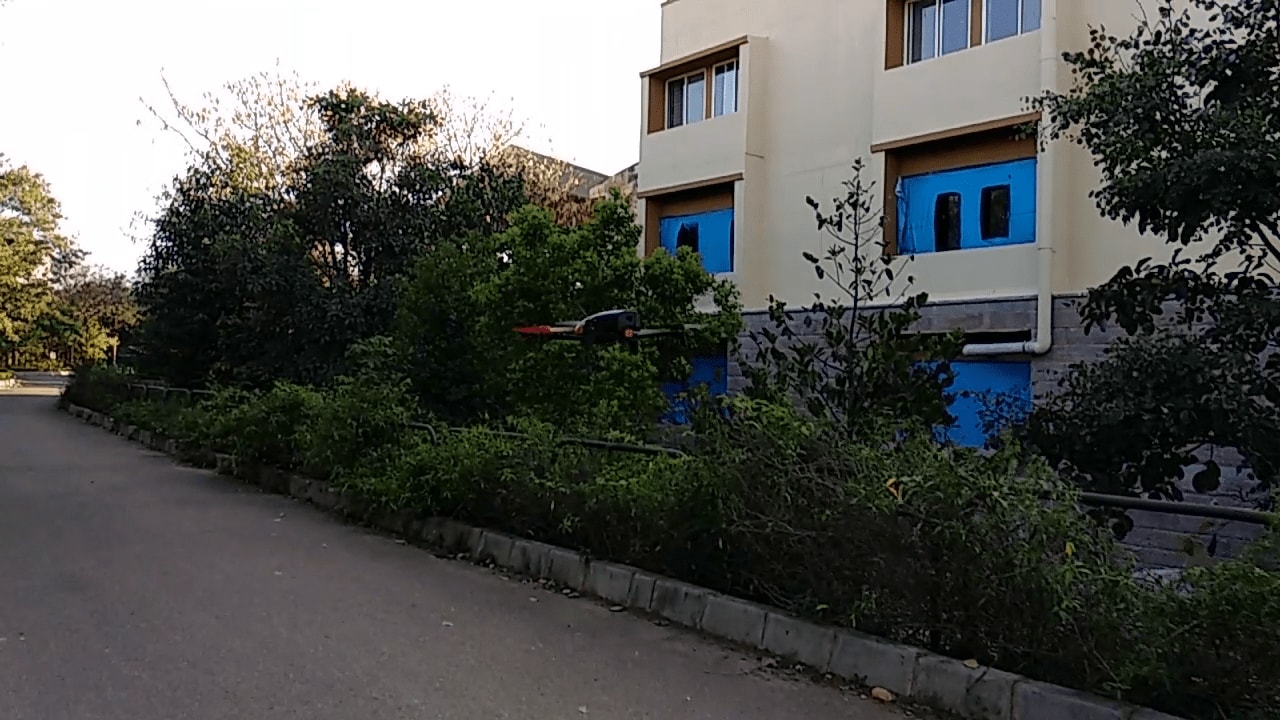} }}
    \subfloat{{\includegraphics[width=2.65cm]{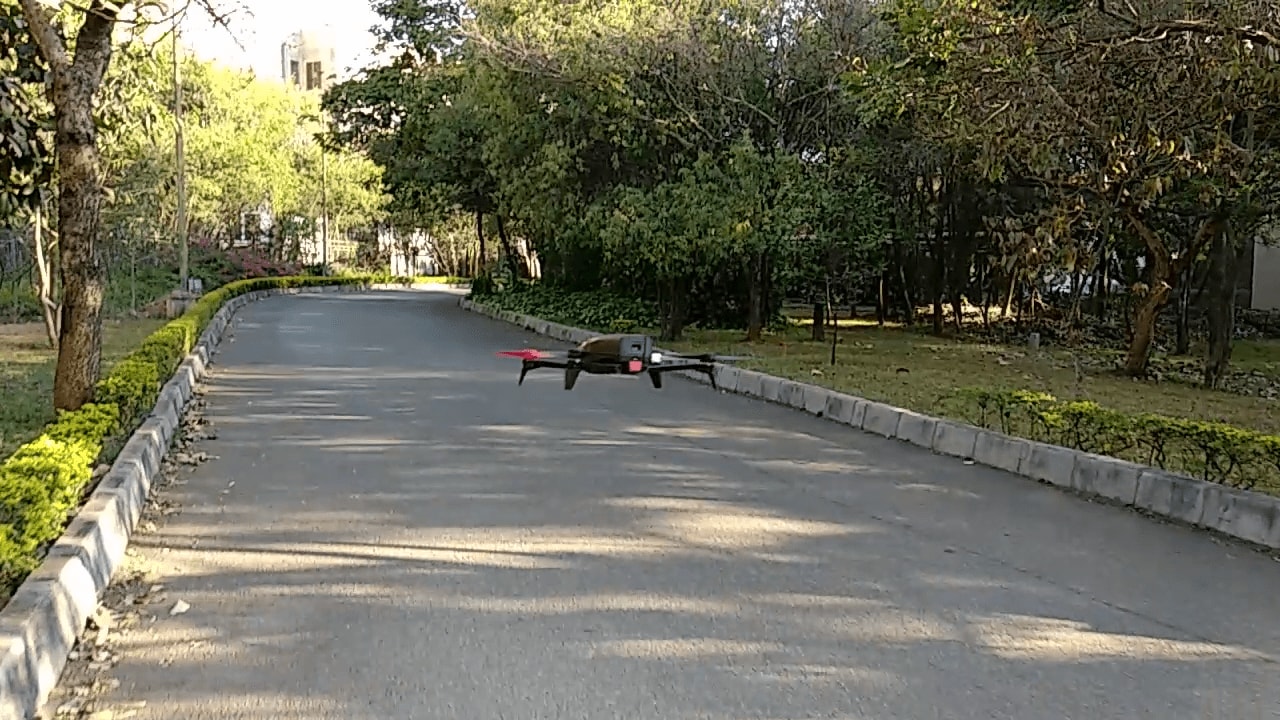} }}
    \subfloat{{\includegraphics[width=2.65cm]{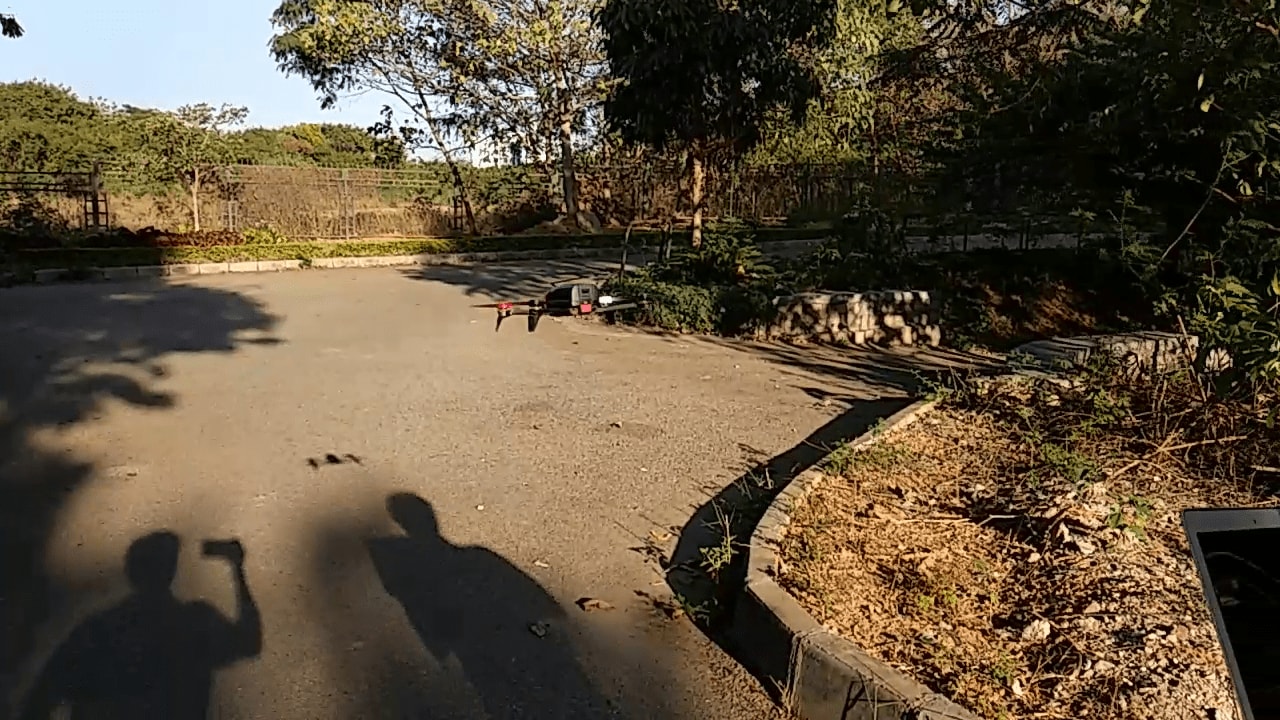} }}
    \quad
    \subfloat{{\includegraphics[width=2.65cm]{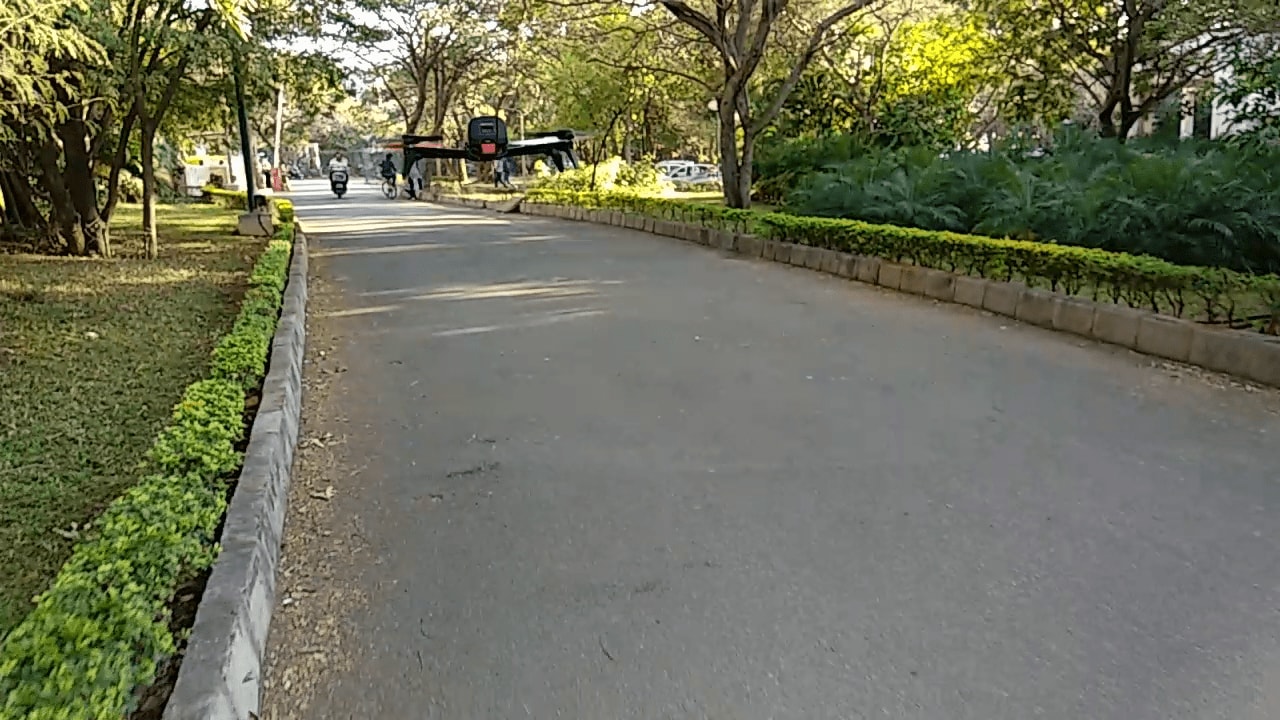} }}
    \subfloat{{\includegraphics[width=2.65cm]{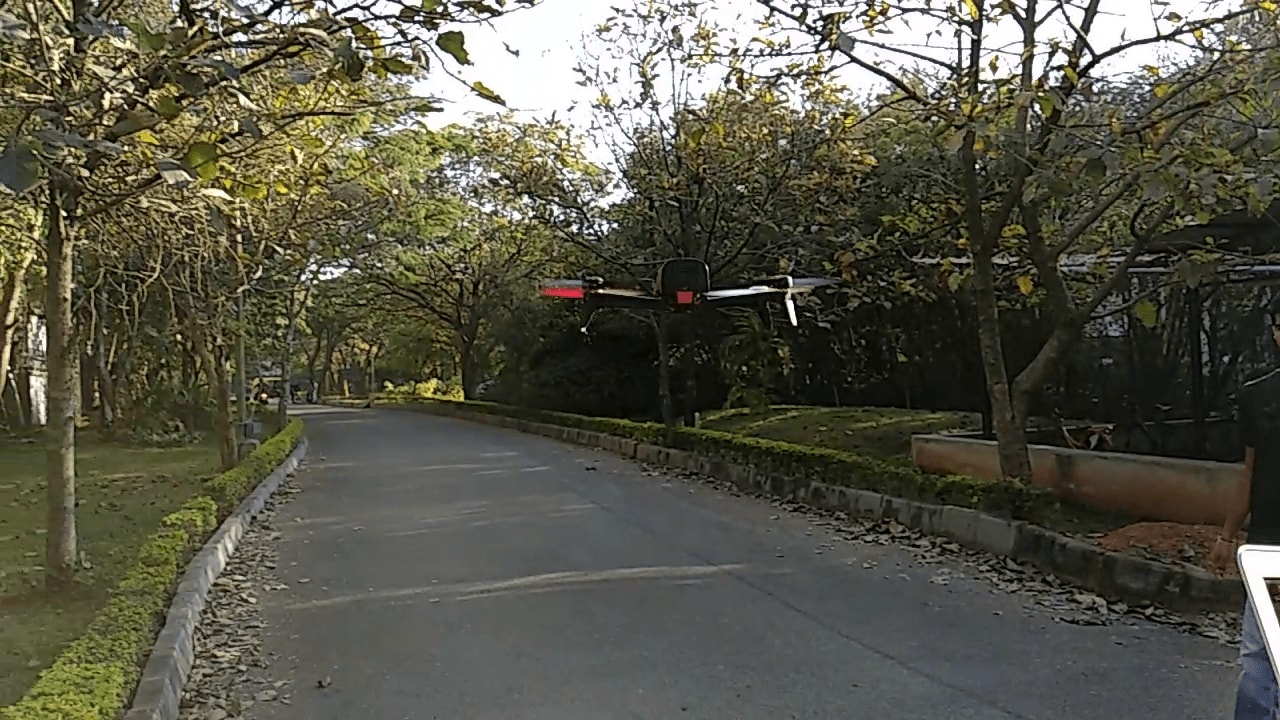} }}
    \subfloat{{\includegraphics[width=2.65cm]{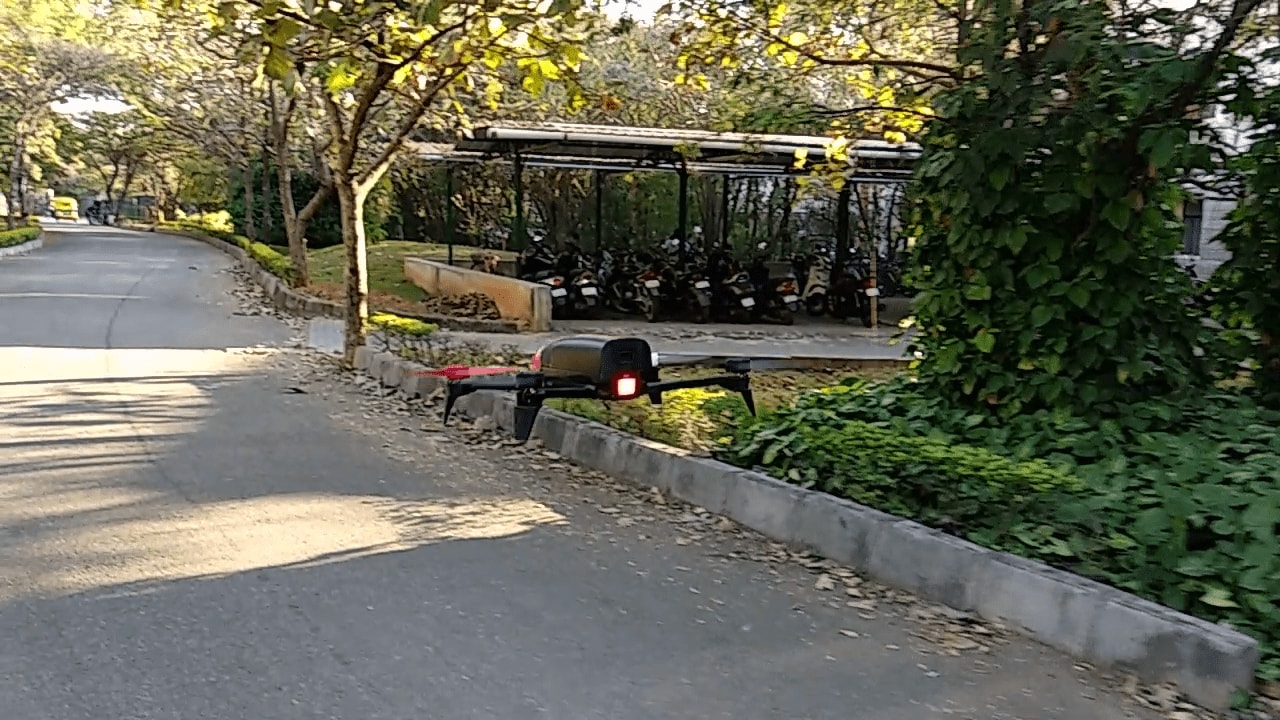} }}

    \caption{Sample images from the testing database of our custom dataset. This is one of the untrained roads where we tested our algorithm. The Bebop tries to isolate the roads from the environments and follow them assisted by junction disambiguation property of MAVNet. The GPS plot for the same stretch in depicted in Fig. 9}
    \label{fig:example}
\end{figure}

\section{Evaluation and Results}

All comparative metrics have been first evaluated on the Udacity Dataset. The images were selected from the custom dataset with T-junctions and multi-junctions to evaluate the accuracy of the junction prediction task. The evaluation metrics as suggested by Loquercio \textit{et. al.} [12] and Ross \textit{et. al.} [11] have been used in this paper for comparison with the state-of-the art techniques. We use Explained Variance (EVA), a metric that is helpful in quantifying the quality of a regressor. It is defined as :

\begin{equation}
\text{EVA} = \frac{Var[Y_{true} - Y_{pred}]}{Var[Y_{true}]}
\end{equation}

Another metric used in this paper is the F-Measure, as suggested by Fritsch \textit{et. al.} [13]. It is a measure of the quality of the classifier used in the system. We use a standard value of 0.9 for $\beta$.

\begin{equation}
\text{F-Measure} = (1 + \beta^2)\frac{PR}{\beta^2P + R}
\end{equation}

Moreover, to evaluate the performance of MAVNet on untrained roads of the custom dataset, we propose to use sample Pearson correlation coefficient, which, for two datasets, is defined as :

\begin{equation}
r = \frac{\sum_{i=1}^n (x_i - \bar{x}) (y_i - \bar{y})}{\sqrt{\sum_{i=1}^n (x_i - \bar{x})^2} \sqrt{\sum_{i=1}^n (y_i - \bar{y})^2}}
\end{equation}

\begin{strip}
  \setlength\arraycolsep{4pt}
  \centering
\scriptsize
  \begin{tabular}{@{}l@{}*{9}{c}c@{}}

    \toprule
    \multirow{2}{*}[-0.6ex]{Architectures} & \multicolumn{9}{c}{{Evaluation Metrics and Performances}} \\
    \cmidrule(lr){3-10}
                         & & F-Measure & Accuracy & RMSE & EVA  & FPS Achieved & Layers & Parameters&\\
    \midrule\addlinespace
    {Random BaseLine}  & &    {0.33$\pm$0.01} & {48.78\%} & {0.4$\pm$0.001} & {0.4$\pm$0.001} & {-} & {-} & {-}\\
    {Inception V3 [18]} & &  {0.927} & {96.53\%} & {0.131} & {0.773} & {23} & {48} & {2.4$\times$10$^7$} \\ 
    {ResNet-50 [14]} & &   {0.925} & {97.15\%} & {0.091} & {0.766} & {9} & {50} & {2.6$\times$10$^7$} \\

    {VGG-16 [15]} & &  {0.852} & {93.14\%} & {0.111} & {0.722} & {12} & {16} & {7.5$\times$10$^6$} \\ 
    {AlexNet [16]} & &  {0.845} & {85.37\%} & {0.353} & {0.778} & {14} & {8} & {6.0$\times$10$^7$} \\ 
    {DroNet [12]} & &   {0.922} & {96.73\%} & {0.098} & {0.721} & {20} & {8} & {3.2$\times$10$^5$} \\
    {\textbf{MAVNet}} & &   {0.945} & {98.44\%} & {0.103} & {0.637} & {30} & {39} & {6.0$\times$10$^6$} \\
    \bottomrule
  \end{tabular}
  \captionof{table}{The table compares the performances of various state-of-the-art methods for autonomus navigation. The metrics of \textbf{EVA} and \textbf{RMSE} have been evaluated on the prediction of the movement commands (pitch and yaw). However, the \textbf{F-1} and \textbf{Accuracy} have been evaluated on the junction prediction task. Hoever, Loquercio \textit{et. al.} [12] generated a probabalistic map of collision detection. In our case, junction disambiguation has been tackled as a binary classification problem, so the F-1 measures have been calculated based on the number of correct hits, correct missed and incorrect hits. The junction disambiguation problem corresponds to the Cityscpaes dataset. As is it could be seen that the MAVNet model performs equally well, if not better. Furthermore, the results mentioned in this table are the outcomes of tests on untrained roads. The computation time for each image was observed to be 0.03225 seconds, which turns out to be almost 30 FPS, if real-time evaluation is considered.} \label{tab:title}
\end{strip}

The metrics for the above evaluations are provided in the image captions of Fig. (6), (7) and (8). The various outputs are individually correlated with the respective training datasets. Moreover, the IMU data plots from expert flights and model flights have been recorded to confirm the accuracy of the MAVNet model. The model takes around 0.03031 seconds, which allows the algorithm to run at a maximum rate of 33 FPS (on custom dataset) as compared to the standard camera input rate of 30 Hz, allowing the drone to achieve a maximum forward velocity of 6 m/sec on straight roads. The straight line policy adopted by Loquercio \textit{et. al.} [12] provided an insight to the maximum distance driven by DroNet, without crashing or going off-track. MAVNet was able to perform a \textbf{357m} continuous autonomous stretch, identifying junctions and taking decisions on-the-go. The GPS plot of the same is shown in Fig. 11.

\begin{figure}[h!]
\hspace*{-0.8cm}
\centering
\includegraphics[width=3.9 in]{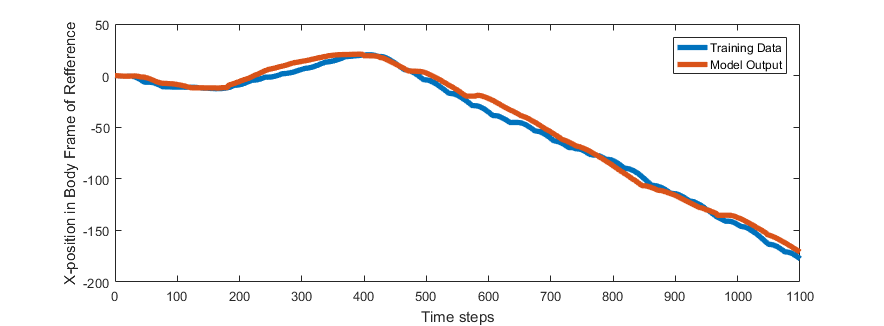}
\caption{Figure indicates the position of the drone along the X-Axis in the body-frame. The position of the drone is recorded by the user on an untrained road (which constitutes the ground truth). MAVNet performs its predictions on the same road and correlation coefficient is calculated to be 0.9977, using Eq. (14).}
\label{fig_chi_dot}
\end{figure}

\begin{figure}[h!]
\hspace*{-0.8cm}
\centering
\includegraphics[width=3.9 in]{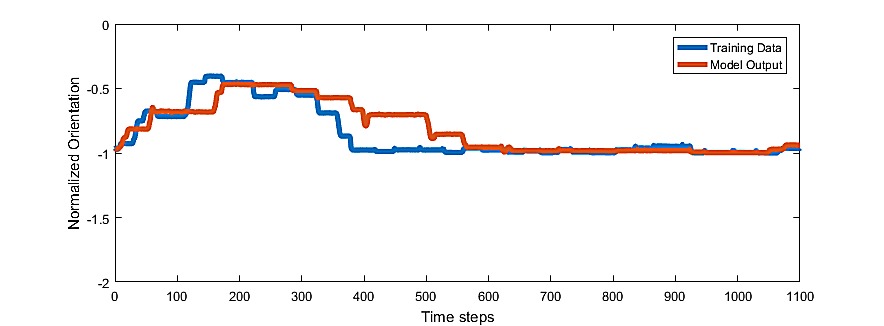}
\caption{Figure indicates the Normalized Orientation of the drone, along the Z-Axis in the body-frame. The degree of Yaw to be executed to make sure that the drone remains in the middle of the road, is recorded by the user on an untrained road (which constitutes the ground truth). MAVNet performs its predictions on the same road and correlation coefficient is calculated to be 0.8534, using Eq. (14).}

\label{fig_chi_dot}
\end{figure}

\begin{figure}[h!]
\hspace*{-0.8cm}
\centering
\includegraphics[width=3.9 in]{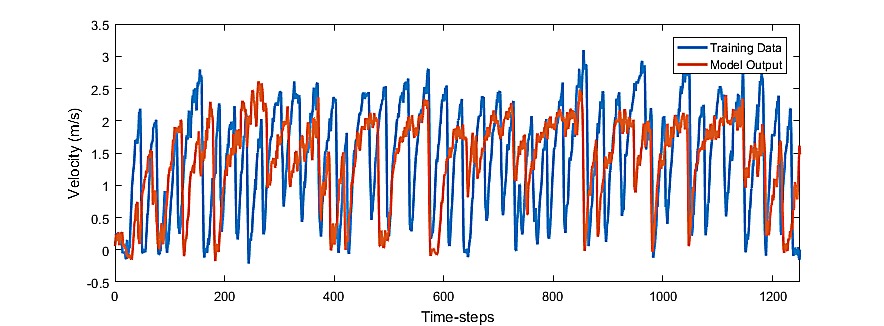}
\caption{Figure indicates the velocity of the drone, along the X-Axis in the body-frame. The maximum speed upto which the drone can move without overshooting the road, is recorded by the user on an untrained road (which constitutes the ground truth). MAVNet performs its predictions on the same road.}
\label{fig_chi_dot}
\end{figure}

\begin{figure}[htp]
\centering
\includegraphics[width=2.7 in]{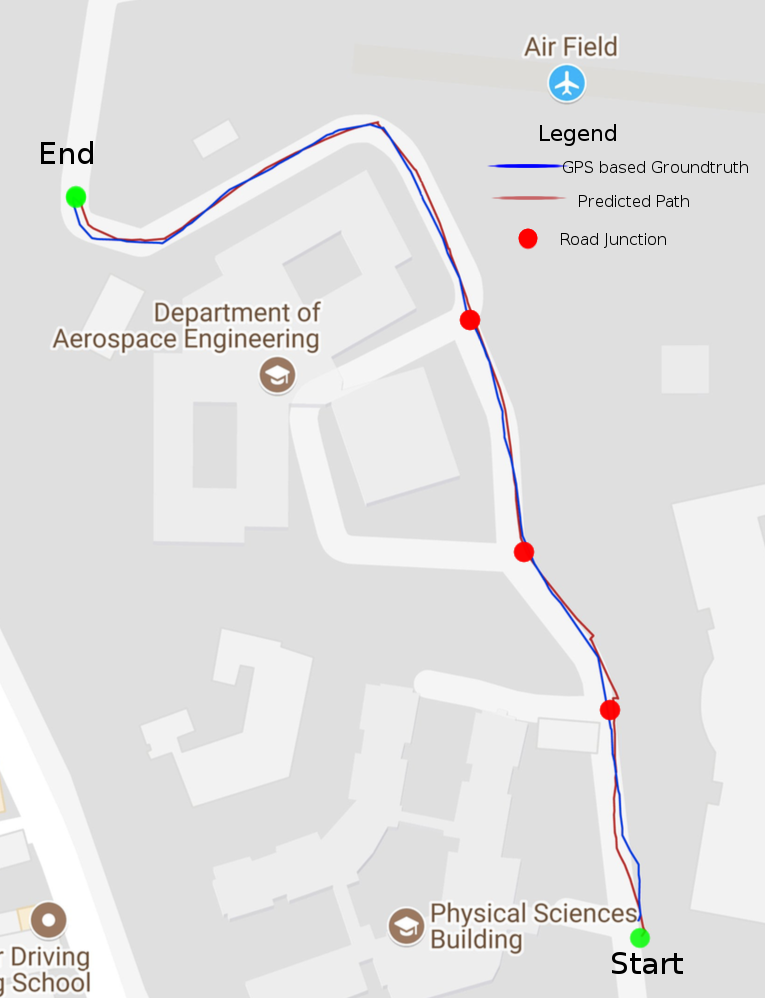}
\caption{Figure indicates the GPS Plots of the paths taken by the UAV. Blue represents the path when the UAV was flown by the expert (which constitutes the ground truth), whereas the red path indicates the path predicted by MAVNet. An interesting thing to note here is that the path shown in the figure does not belong to the training dataset and the total stretch of this patch is approximately is 357 m.}
\label{fig_chi_dot}
\end{figure}

\section{Supplementary Material}
The trained model and the final codes for data collection from every simulation environment have been made available at the following GitHub repository : 
\href{$https://github.com/sudakshin/imitation_learning$}{$https://github.com/sudakshin/imitation_learning$}

\section{Discussions and Conclusion}
Although we have not integrated GPS positioning with our current architecture GPS fusion could be done to tag the road-junctions. In case they are missed by MAVNet, the program could read the tag information and execute the turning. However, the fundamental problem associated with GPS/GNSS signalling systems is that the average power of the signal is around $10^{-7}$ W. Even though the GPS receivers have powerful amplifiers but the quality of signals that we receive actually also depends upon the geometric orientations of the satellites, signal blockage, atmospheric parameters, and receiver design features. The usual range of GPS for typical smart-phone receivers is around 4-5m, which directly results in offsets. This was the primary rationale behind keeping our architecture free from the advantages and disadvantages of GPS.

Anyhow, since the proposed architecture is close to and end-to-end system for autonomous road navigation, there are still some issues where improvements could be made. For instance, the performance of our algorithm drops to 81\% when there are sharp shadows involved. Sharp shadows, as mentioned earlier, add edges to the image frame, which makes it difficult for the algorithm to identify the roads clearly. There has been some research on shadow removal from images, but those were primarily focused on a single continuous patch of shadow, and not an incoherent one. Another case is where there are patched roads with sharp turns. In such cases, we have observed that our system performance degrades a bit. However, even with some shortcomings, we hope to have come up with a novel solution of learning based autonomous navigation, an idea which we hope to get implemented extensively. This system is not claimed to be complete in itself but if used in conjuction with obstacle avoidance and localization modules, there is a great potential in its implementation not only in the case of drone navigation but also can be extended to autonomous passenger transports.

Further, there is a great scope for improving the accuracy and reliability of the MAVNet by adding a temporal characteristic to the existing spatial nature of the algorithm by making use of Long-term Recurrent Convolutional Networks (LRCNs).

\end{document}